%% file: main.tex
\documentclass[preprint,5p]{elsarticle} 

\usepackage{amssymb}
\usepackage{amsmath}

\usepackage[final]{changes}

\journal{Solar Energy}

\input{preamble.tex}

\begin{document}

\begin{frontmatter}



\title{Forecasting Solar Energy Using a Single Image}


\author[1]{Jeremy Klotz\corref{cor1}} 
\ead{jklotz@cs.columbia.edu}
\cortext[cor1]{Corresponding author}

\author[1]{Shree K. Nayar} 
\ead{nayar@cs.columbia.edu}

\affiliation[1]{organization={Computer Science Department, Columbia University},
            addressline={500 West 120th St.}, 
            city={New York},
            postcode={10027}, 
            state={NY},
            country={USA}}

\begin{abstract}
\input{sections/abstract.tex}
\end{abstract}




\end{frontmatter}


\input{sections/sec01_introduction.tex}
\input{sections/sec02_camera_orientation.tex}
\input{sections/sec03_sun_sky.tex}
\input{sections/sec04_scene_irradiance.tex}
\input{sections/sec05_experiments.tex}
\input{sections/sec06_citibike.tex}
\input{sections/sec07_solaris.tex}
\input{sections/sec08_discussion.tex}

\input{sections/acknowledgements.tex}

\section*{Declaration of generative AI and AI-assisted technologies in the manuscript preparation process.}
During the preparation of this work the authors used a large language model in order to improve the readability of the paper. After using this tool/service, the authors reviewed and edited the content as needed and take full responsibility for the content of the published article.

\appendix

\renewcommand{\thesection}{\Alph{section}}

\input{sections/app01_pose.tex}

\input{sections/app07_sky_model.tex}

\input{sections/app02_scene_irradiance.tex}

\input{sections/app04_experiment_details.tex}

\input{sections/app03_extended_results.tex}

\input{sections/app06_panel_orientation.tex}

\bibliographystyle{elsarticle-num} 
\bibliography{Solar.bib}



\end{document}

%% file: preamble.tex
\usepackage{graphicx}%
\usepackage{multirow}%
\usepackage{amsmath,amssymb,amsfonts}%
\usepackage{amsthm}%
\usepackage{mathrsfs}%
\usepackage[title]{appendix}%
\usepackage{xcolor}%
\usepackage{textcomp}%
\usepackage{manyfoot}%
\usepackage{booktabs}%
\usepackage{algorithm}%
\usepackage{algorithmicx}%
\usepackage{algpseudocode}%
\usepackage{listings}%
\usepackage{xspace}
\usepackage{siunitx}
\usepackage{comment}
\usepackage{mathtools}
\usepackage{hyperref}
\usepackage{cleveref}
\usepackage{makecell}
\usepackage{threeparttable}

\newcommand{\vs}{\vec{s}}
\newcommand{\vn}{\vec{n}}
\newcommand{\dw}{d\Omega}
\newcommand{\Lpatch}{L(\vs)}
\newcommand{\Epatch}{E_{patch \rightarrow panel}}
\newcommand{\Escene}{E_{scene}(t)}
\newcommand{\wsunlit}{\Omega_{\text{sunlit}}}
\newcommand{\wshadowed}{\Omega_{\text{shadowed}}}
\newcommand{\wboundary}{\Omega_{\text{boundary}}}

\makeatletter
\DeclareRobustCommand\onedot{%
  \futurelet\@let@token\@onedot
}
\def\@onedot{%
  \ifx\@let@token.\else.\null\fi\xspace
}
\def\etal{\emph{et al}\onedot}
\makeatother

\makeatletter
\def\ps@pprintTitle{%
 \let\@oddhead\@empty
 \let\@evenhead\@empty
 \let\@oddfoot\@empty
 \let\@evenfoot\@empty
}
\makeatother

%% file: sections/abstract.tex
Solar panels are increasingly deployed in cities on rooftops, walls, and urban infrastructure. Although the panel costs have fallen in recent years, the soft costs of installing them have not. These soft costs include assessing the illumination (irradiance) of a panel, which is typically performed using a 3D model that fails to capture small nearby structures that impact the irradiance. Our approach uses a single image taken at the panel's location to forecast its irradiance at any time in the future. We use visual cues in the image to find the camera's orientation and the portion of the sky visible to the panel in order to forecast the irradiance due to the sun and the sky. In addition, we show that the irradiance due to reflections from nearby buildings varies smoothly over time and can be forecasted from the image. This approach enables assessing the solar energy potential of any surface and forecasting the temporal variation of a panel's irradiance. We validate our approach using real irradiance measurements in urban canyons. 
We show that our approach often yields more accurate irradiance forecasts compared to conventional irradiance-based transposition methods and 3D model-based simulations.
We also show that a single spherical image can be used to find the best fixed orientation of a panel. Finally, we present \textit{Solaris}, a device to capture the image seen by a panel in a variety of urban settings.

%% file: sections/sec01_introduction.tex
\section{Introduction}

Solar panels are becoming ubiquitous in urban settings: panels are widely installed on city rooftops, walls, and infrastructure, and creative solutions are emerging to integrate panels into urban materials~\cite{kuhnReviewTechnologicalDesign2021, shuklaComprehensiveReviewDesign2016}. For example, transparent panels can serve as windows~\cite{traverseEmergenceHighlyTransparent2017, solarwindowtechnologiesLiquidElectricity, nextenergytechnologiesNEXTOPVTransparent}, and panels with the appearance of building materials can be used as walls~\cite{efacadeMitrex}. As the demand for electricity in urban environments is projected to grow~\cite{newyorkindependentsystemoperator2024ReliabilityNeeds24}, urban solar panels provide one path toward reducing a city's demand on the grid.

While the cost to manufacture solar panels has decreased by roughly $5\times$ in the last ten years~\cite{nationalrenewableenergylaboratorynrelSolarManufacturingCost}, the soft costs of installing them have not~\cite{nationalrenewableenergylaboratorynrelSolarInstalledSystem, oshaughnessyAddressingSoftCost2019}. 
These soft costs include site assessment, design, permitting, inspection, interconnection, and labor, and they now account for $50\%$ of the total system cost in the United States.
For example, in 2024, the soft costs of a typical commercial solar panel installation were $0.78\,\$/\text{Wdc}$ out of a total system cost of $1.6\,\$/\text{Wdc}$~\cite{nationalrenewableenergylaboratorynrelSolarInstalledSystem}. Given these large soft costs, there is a clear need to understand the return on investment before installing a solar panel.

The first step in determining a panel's return on investment (i.e.,~estimating how much energy it will produce) is to forecast the panel's illumination at any time in the future. Unlike panels in an open field, panels in urban settings are often installed in urban canyons formed by surrounding buildings that obstruct large parts of the sky. A panel in such a canyon is illuminated by only a portion of the sky and is exposed to a variety of complex time-varying effects such as shadows and reflections due to nearby buildings. If a panel is mounted on an actuator, sensors can be used to adjust the orientation of the panel to maximize its illumination (irradiance)~\cite{klotzMinimalSensingOrienting2025}. However, most panels in urban environments are simply attached to a surface and hence are fixed in orientation. In these cases, the irradiance received by the panel, and hence the energy harvested by it, depends heavily on the properties of the urban canyon in which it resides. As a result, an accurate assessment of the solar energy potential of an urban surface must consider the effects of the surrounding environment.

Given the location and orientation of a panel, our goal is to develop a practical method for predicting its irradiance at any time in the future. Such a method has three implications: 
(a)~Before installing a panel on a surface, we can assess the surface's solar energy potential to determine the return on investment.
(b)~For already installed panels, we can forecast the temporal variation in the harvested energy, which can be used to forecast demand and optimize generation on the grid.
(c)~If the orientation of a panel can be mechanically adjusted (e.g.,~a panel mounted on a pole), we can determine the panel orientation that would yield the maximum harvested energy. 

Existing methods forecast the irradiance of a panel by simulating the illumination seen by the panel using an explicit 3D model of the environment~\cite{jakubiecMethodPredictingCitywide2013, freitasModellingSolarPotential2015, mardaljevicIrradiationMappingComplex2003, zhuEffectUrbanMorphology2020, robinsonSolarRadiationModelling2004, robinsonSUNtoolNewModelling2007, compagnonSolarDaylightAvailability2004, jakicaStateoftheartReviewSolar2018, kosmopoulosRayTracingModelingUrban2024, anSolarEnergyPotential2023, googleProjectSunroof, martinez-rubioEvaluatingSolarIrradiance2016, chengSolarEnergyPotential2020, andresTimevaryingRayTracing2023}. While this approach can simulate the time-varying illumination phenomena mentioned above, it requires a detailed 3D model of the panel's environment, which is typically acquired using 3D scanning (such as LiDAR) or estimated using satellite imagery. These methods of acquiring a 3D model, however, only provide a crude representation of an urban canyon---they cannot capture the fine-grained structures near a panel that obstruct the panel's view of the sky or the environment. This is generally not an issue for utility-scale panel installations in open fields. However, fine-grained structures are inescapable in urban environments. 

Consider a panel on a city rooftop. Virtually every roof in an urban environment is sprinkled with tiny structures such as parapets, railings, pipes, and vents that existing 3D models fail to capture. Even though these structures are small, their impact on a panel's view of the sky is significant when the panel is close by. In fact, a small structure right next to a panel has the same impact on the panel's view of the sky as that of a distant building. As we will show, existing methods for forecasting a panel's irradiance using a 3D model yield inaccurate results by not accounting for the effects of tiny structures nearby. Our approach accounts for the effects of such structures by using a single image taken from a panel's location, which captures details of every structure visible to the panel, ranging from intricate, nearby structures to large, distant buildings. In our case, we do not explicitly capture the 3D structure or material properties of the surrounding environment. Instead, we show that a single image reveals both how nearby structures obstruct a panel's view of the sky and how light reflected by nearby buildings interacts with the panel.

The irradiance of a panel in an urban canyon depends on the portion of the sky visible to it, which we refer to as the ``sky aperture.'' Calcabrini \etal~\cite{calcabriniSimplifiedSkylinebasedMethod2019} proposed a method to forecast the annual irradiance of a panel using two metrics: the sky view factor and the sun coverage factor. While this method is computationally simple, it requires precise knowledge of the sky aperture, which must be measured using a device that is accurately leveled and oriented~\cite{meteonormHoricatcher} or estimated from a 3D model of the environment. As we will show, existing 3D models do not capture the small urban structures near a panel, yielding inaccurate estimates of a panel's sky aperture. Our goal is to develop a practical approach to forecast a panel's irradiance without any prior knowledge of the surrounding environment and without using any device that requires careful alignment. Prior work has also used a camera placed near the panel to capture a video of the illumination seen by the panel to forecast its irradiance~\cite{zhangDeepPhotovoltaicNowcasting2018, julianComputationalImagingLongTerm2025}. This approach, however, requires a camera to be installed near a panel to capture a video over a long time duration, which is infeasible in many urban scenarios.

\begin{figure*}[t]
    \centering
    \includegraphics[]{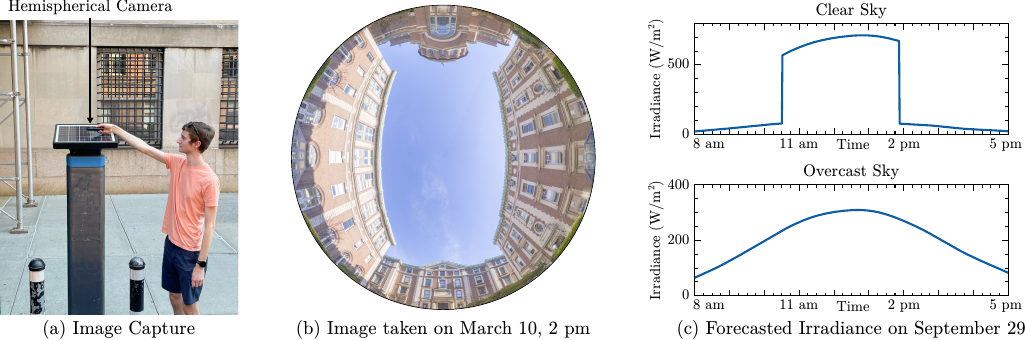}
    \caption{
        \textbf{What does a single image reveal about the future?}
        (a)-(b)~An individual walks up to a solar panel (a) and takes a single hemispherical image on March 10 (b). 
        We use this image to forecast the panel irradiance at any time in the future for given sky conditions. Visual cues in the image are used to automatically determine the orientation of the camera and the portion of the sky visible to the panel. In addition, even though we have no prior knowledge of the environment's 3D structure, the image reveals how shadows and reflections due to nearby buildings interact with the panel. We leverage all of this visual information embedded in the image to forecast the irradiance of the panel.
        (c)~The forecasted panel irradiance as a function of time on September 29---six months after the image in (b) was taken---for two different sky conditions. 
        This approach to forecast a panel's irradiance from a single image can be used both to assess the solar energy potential of an urban surface and to forecast the temporal variation in a panel's irradiance (and hence the energy harvested by it), which is valuable information for the electric grid.
    }
    \label{fig:intro}
\end{figure*}

Our approach forecasts the irradiance of a solar panel using a single image without any knowledge of the camera's orientation or the surrounding environment. As an example, suppose an individual walks up to the panel in \cref{fig:intro}(a) and captures the hemispherical (fisheye) image in \cref{fig:intro}(b). The shadows and scene lines in the image provide visual cues that can be used to automatically determine the camera's orientation in the Earth coordinate frame. In addition, we can determine the panel's sky aperture by segmenting the sky from the image. This allows us to forecast how the portion of the sky visible to the panel changes over time. We also show that the image reveals the geometry and material properties of the canyon in which the panel resides. As a result, we can forecast the panel's irradiance due to reflections from the surrounding environment, which we refer to as ``scene irradiance.'' By leveraging all of this visual information embedded in the image, we can forecast the future irradiance of the panel for any given sky condition. \Cref{fig:intro}(c) shows the forecasted irradiance as a function of time on September 29---six months after the image in \cref{fig:intro}(b) was taken---for a clear sky and an overcast sky. 
Given that our method uses a single image without any prior knowledge of the surrounding environment, it allows for the rapid assessment of the solar energy potential of any urban surface. Once the panel is installed, our approach can be used to forecast the panel irradiance as a function of time (\cref{fig:intro}(c)), which is valuable information for the electric grid~\cite{antonanzasReviewPhotovoltaicPower2016}.

We validate our approach using real irradiance measurements made in four urban canyons under clear, partly cloudy, and overcast skies. 
We compare the performance of our approach with two conventional methods for forecasting irradiance: a widely used irradiance-based transposition model~\cite{gilmanSAMPhotovoltaicModel2018} and a 3D model-based simulation. Compared to these alternative methods, our approach often yields more accurate forecasts of a panel's irradiance in the four urban environments.
In addition, we show that by capturing a single spherical image, we can find the best fixed panel orientation that would yield the maximum annual irradiance. We demonstrate this capability by capturing a spherical image at four already installed solar panels in Manhattan. We determine how the orientation of each of these panels should be adjusted to maximize the annual irradiance. Finally, we present \textit{Solaris}, a device for capturing the images our method requires in a variety of urban settings. 

%% file: sections/sec02_camera_orientation.tex
\section{Estimating Camera Orientation}
\begin{figure*}[t]
    \centering
    \includegraphics[]{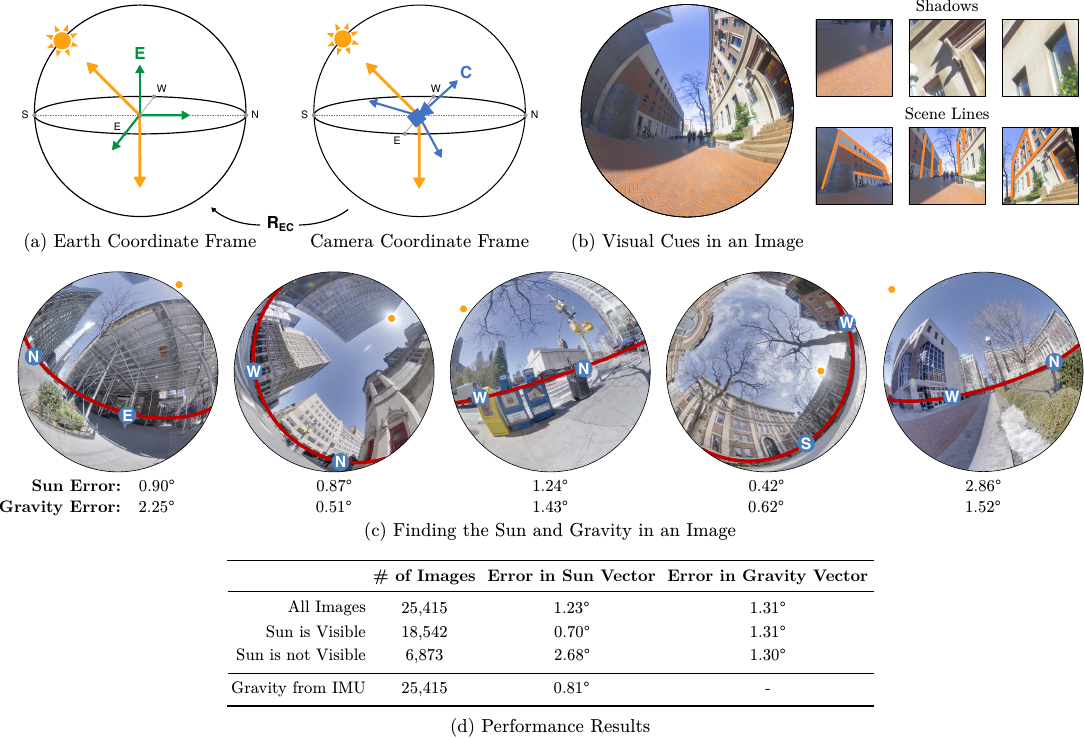}
    \caption{
        \textbf{Estimating camera orientation using a single image.}
        (a)~The Earth coordinate frame $E$ and camera coordinate frame $C$ are related by a rotation $R_{EC}$. We determine $R_{EC}$ by finding the directions of the sun and gravity (orange vectors) in both coordinate frames. The direction of the sun in the Earth coordinate frame is computed using the date, time, and GPS location of the image. 
        (b)~Visual cues in the image reveal the directions of the sun and gravity in the camera coordinate frame. The inset perspective images show shadows cast by buildings and small objects that reveal the sun's direction (top row) and scene lines that reveal the direction of gravity (bottom row). We trained a neural network that takes a single hemispherical image as input and estimates the direction of the sun and gravity in the camera coordinate frame. 
        (c)~The output of the neural network for five different images, with the prediction error listed below each image. The orange dot indicates the predicted sun direction, and the red line indicates the predicted horizon line, which is orthogonal to the gravity vector. Note that the sun need not lie within the field of view of the captured image. The only requirement is that the image contains shadows that encode the sun's direction. 
        (d)~The average error in the predicted sun and gravity vectors computed over 25,415 test images. The network accurately predicts the direction of the sun and gravity in diverse urban settings. 
    }
    \label{fig:camera-orientation}
\end{figure*}

In order to forecast a panel's irradiance from a single image, we first need to determine the camera's orientation in the Earth coordinate frame. As shown in \cref{fig:camera-orientation}(a), the Earth and camera coordinate frames are related by a rotation $R_{EC}$. One approach to determine $R_{EC}$ is to use an inertial measurement unit (IMU) attached to the camera, which uses an accelerometer to measure the direction of gravity and a magnetometer to measure the direction of magnetic north. In ideal conditions, these measurements are sufficient to determine $R_{EC}$. However, in urban settings, nearby buildings and objects disturb the magnetic field, and hence the magnetometer readings are unreliable.

Since magnetic north cannot be reliably measured by an IMU in urban settings, our approach uses the direction of the sun and gravity to determine the camera orientation. The sun vector in the Earth coordinate frame can be computed using the date, time, and GPS location corresponding to the captured image~\cite{michalskyAstronomicalAlmanacAlgorithm1988,meeusAstronomicalAlgorithms1991}. Therefore, determining the camera orientation reduces to finding the sun and gravity vectors in the camera coordinate frame.

Consider the image in \cref{fig:camera-orientation}(b). Although the sun is not visible in the image, shadows cast by buildings and smaller objects provide visual cues that indicate the sun's direction (see the inset images in the top row of \cref{fig:camera-orientation}(b))~\cite{lalondeEstimatingNaturalIllumination2012}. Similarly, vertical and horizontal lines in the scene reveal the direction of gravity (see the inset perspective images computed from the captured hemispherical image in the bottom row of \cref{fig:camera-orientation}(b))~\cite{workmanHorizonLinesWild2016, xianUprightNetGeometryAwareCamera2019, hold-geoffroyPerceptualMeasureDeep2018}. 

Based on this observation, we trained a neural network to find the direction of the sun and gravity in a hemispherical image taken in an urban environment.  The network consists of a TinyViT backbone~\cite{wuTinyViTFastPretraining2022}, followed by two separate multi-layer perceptrons (MLPs) to predict the directions of the sun and gravity vectors in the camera coordinate frame. 
Each MLP has two hidden layers that are 256 units wide with ReLU activation functions. Similar to prior work~\cite{workmanHorizonLinesWild2016, hold-geoffroyPerceptualMeasureDeep2018}, the outputs of each MLP are probability distributions over the azimuth and zenith angles of each vector, discretized into $1\unit{\degree}$ bins. 
The input to the network is a high-dynamic range (HDR) $512\times512\,\unit{\text{px}}$ image.
We trained the network using a dataset of one million rendered HDR hemispherical images in a simulated urban environment. We then fine-tuned the network on real HDR hemispherical images taken in urban environments from the \textit{UrbanSky} dataset~\cite{klotzMinimalSensingOrienting2025}. Once we have determined the sun and gravity vectors in both coordinate frames, we use the Kabsch algorithm to compute the rotation $R_{EC}$ between the coordinate frames. Further details of the network training procedure are provided in \cref{app:pose}.

\Cref{fig:camera-orientation}(c) shows the camera orientation estimated by the neural network for five different images. The orange dot shows the predicted sun direction, and the red line shows the predicted horizon line, which is orthogonal to the predicted gravity vector.  Notice that the sun does not need to be visible in the image; the only requirement is that the sun be unobstructed by clouds so that the image contains shadows that reveal the sun's direction. In \cref{fig:camera-orientation}(c), we visualize the camera's orientation in the Earth coordinate frame by overlaying the cardinal directions (e.g.,~north and south) on each image. 

The neural network's performance across 25,415 test images from the \textit{UrbanSky} dataset is summarized in \cref{fig:camera-orientation}(d). The average error in the predicted sun vector is $1.23\unit{\degree}$, and the average error in the predicted gravity vector is $1.31\unit{\degree}$. If the camera includes an IMU, we can pass the gravity vector measured by the IMU as an additional input to the MLP that predicts the direction of the sun, which reduces the average error in the predicted sun vector to $0.81\unit{\degree}$ (last row in \cref{fig:camera-orientation}(d)). These results show that our approach accurately finds the orientation of the camera in diverse urban environments, even when the sun is not visible in the image. 
Please see \cref{app:pose} for additional results and analysis of the network's performance.

%% file: sections/sec03_sun_sky.tex
\section{Forecasting Irradiance Due to the Sun and Sky}
Now that we can find the camera's orientation in the Earth coordinate frame, we will forecast the panel irradiance due to the sun and the sky by modeling how the sky changes with time. 

\begin{figure*}[t]
    \centering
    \includegraphics[]{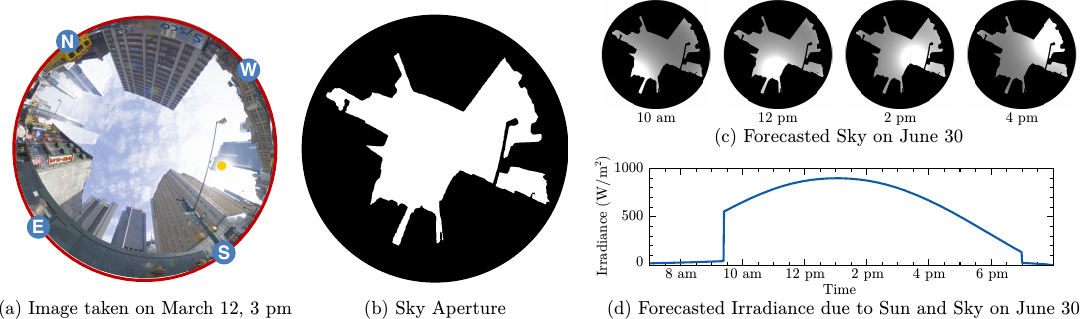}
    \caption{
        \textbf{Forecasting irradiance due to the sun and the sky.}
        (a)~A hemispherical image taken on March 12 at 3pm, with the estimated camera orientation overlaid on top.
        (b)~We use an image segmentation algorithm to find the sky aperture.
        (c)~Using the sky aperture and the estimated camera orientation, we can forecast how the illumination from the sky changes over time for any given sky condition. Here we show the forecasted illumination on June 30 for a clear sky.
        (d)~The forecasted panel irradiance due to the sun and the sky on June 30. This prediction can be used to forecast ramp events in the harvested energy (caused by the sun moving in and out of the sky aperture), which affect the net demand on the grid.
    }
    \label{fig:sky-sun}
\end{figure*}

Consider the image in \cref{fig:sky-sun}(a) taken at 3 pm on March 12. We use Segment Anything~\cite{kirillovSegmentAnything2023}, an image segmentation algorithm, to identify the sky aperture seen by the panel (\cref{fig:sky-sun}(b)). We model the illumination within the sky aperture using the Perez sky model~\cite{perezAllweatherModelSky1993}, which parameterizes the sky conditions using two irradiance measurements: the diffuse horizontal irradiance (DHI) and direct normal irradiance (DNI). These irradiances can either be measured by a nearby weather station or estimated using satellite imagery.\footnote{In cases in which we are only interested in forecasting the annual irradiance of a panel rather than the instantaneous irradiance, the typical meteorological year at the panel's location provides the requisite irradiance measurements.} 
Together with the estimated camera orientation, the sky aperture and Perez sky model allow us to predict how the panel's illumination from the sky changes with time for any given sky conditions. \Cref{fig:sky-sun}(c) shows the forecasted sky illumination on June 30---three months after the image in \cref{fig:sky-sun}(a) was taken---for clear sky conditions. 

We use the forecasted illumination to compute the panel irradiance due to the sky as an integral over the entire sky,
\begin{equation}
    E_{sky} = \int\limits_{0}^{2\pi} \int\limits_{0}^{\pi} L_{sky}(\theta, \phi) \, A(\theta, \phi) \, \sin\theta \, d\theta \, d\phi, \label{eq:Esky}
\end{equation}
where $L_{sky}(\theta, \phi)$ is the radiance from the sky computed using the Perez sky model (see \cref{app:sky-model} for details), $A(\theta, \phi)$ is a binary function representing the sky aperture, and $\theta$ and $\phi$ are the zenith and azimuth angles in the Earth coordinate frame, respectively. We model the sun as a point source that is only visible to the panel when the sun is within the sky aperture. Thus, the irradiance due to the sun is computed as 
\begin{equation}
    E_{sun} = E_{DNI} \, A(\theta_s, \phi_s)  \, (\vn \cdot \vs), \label{eq:Esun}
\end{equation}
where $E_{DNI}$ is the direct normal irradiance, $\vn$ is the normal vector of the panel, $\vs$ is the direction of the sun from the panel, and $\theta_s$ and $\phi_s$ are the sun's zenith and azimuth angles, respectively. \Cref{fig:sky-sun}(d) shows the forecasted panel irradiance due to the sun and the sky on June 30. 
In this example, the sun is visible to the panel at starting at 9:30 am when it appears above the buildings, causing the panel irradiance and hence the harvested energy to increase dramatically. This temporal forecast of the irradiance due to the sun and the sky is valuable information for the grid as it shows the ramp events in the panel's harvested energy, which affect the net demand on the grid~\cite{antonanzasReviewPhotovoltaicPower2016}.

Note that this approach can model the illumination from the sun and the sky for any spatially smooth sky conditions, ranging from a clear sky to an overcast sky~\cite{perezAllweatherModelSky1993}. However, it cannot model the transient effects of small, individual clouds that briefly obstruct the sun, since the motion of a single cloud is a random process that is notoriously difficult to forecast.

%% file: sections/sec04_scene_irradiance.tex
\section{Forecasting Irradiance Due to the Scene}
\label{sec:scene-irradiance}
So far, we have only considered the irradiance of the panel due to the sun and the sky. In urban settings, however, the panel is also illuminated by nearby buildings. Via large-scale simulations we have found that the irradiance contribution due to the scene accounts for $12\%$, on average, of the total annual irradiance received by a solar panel and hence cannot be ignored. At first glance, forecasting the panel irradiance due to the scene (we refer to this as scene irradiance) from a single image appears to be infeasible. We will show that even though the underlying illumination phenomena are complex, the scene irradiance is invariant to the scale of the scene and is a smooth function with respect to time.

\subsection{Contribution of Irradiance Due to the Scene}
\begin{figure*}[t]
    \centering
    \includegraphics[]{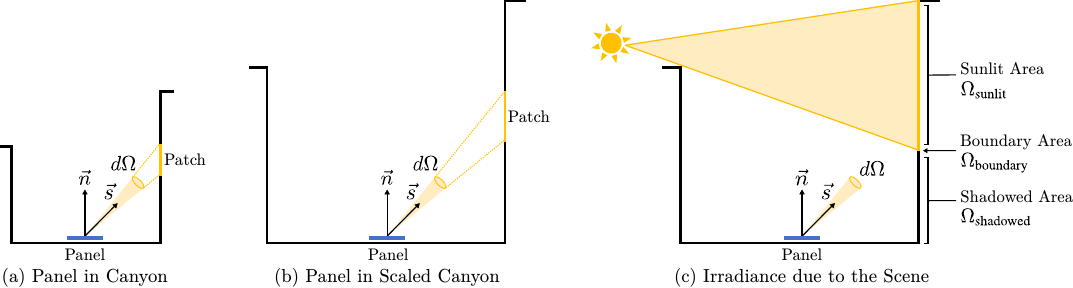}
    \caption{
        \textbf{Properties of scene irradiance.}
        (a)~A solar panel in an urban canyon. The irradiance of the panel due to the yellow scene patch depends on the radiance of the patch along  direction $\vs$ and the solid angle $\dw$ subtended by the patch.
        (b)~The solar panel in a scaled version of the canyon in~(a). Note that the yellow  patch's radiance and the solid angle $\dw$ subtended by it are unchanged. As a result, the panel irradiance due to the patch, and hence the entire scene, is invariant to scene's scale.
        (c)~Suppose the canyon is illuminated by the sun moving through the sky. If we assume the scene to be Lambertian, then the irradiance of the panel due to the patches in the sunlit area ($\wsunlit$) and the shadowed area ($\wshadowed$) changes slowly with time. Although the irradiance of the panel due to the patches on the boundary of the shadow $(\wboundary$) changes dramatically as the shadow moves across the boundary area, the total solid angle subtended by boundary area is small compared to the solid angle subtended by the sunlit and shadowed areas. As a result, the contribution of the boundary area to the scene irradiance is negligible, and hence the scene irradiance varies smoothly with time.
    }
    \label{fig:scene-irradiance-props}
\end{figure*}

Consider the solar panel in an urban canyon in \cref{fig:scene-irradiance-props}(a). The irradiance of the panel due to the yellow scene patch is
\begin{equation}
    \Epatch = \Lpatch \, (\vn \cdot \vs) \, \dw,
    \label{eq:scale-invariance}
\end{equation}
where $\Lpatch$ is the radiance of the patch in the direction $\vs$, $\vn$ is the normal vector of the panel, and $\dw$ is the solid angle subtended by the patch (see \cref{fig:scene-irradiance-props}(a)). If we scale the entire scene as illustrated in \cref{fig:scene-irradiance-props}(b), then the radiance $\Lpatch$ from the patch does not change since the patch's orientation and albedo remain the same. In addition, even though the scaled patch in \cref{fig:scene-irradiance-props}(b) is larger, it subtends the same solid angle $\dw$. As a result, $\Epatch$ is unchanged in \cref{fig:scene-irradiance-props}(b). By extending this argument to the entire scene, which can be represented as a large collection of patches, we see that the scene irradiance is invariant to the scene's scale. In short, it does not matter if the solar panel resides in a small canyon (as in \cref{fig:scene-irradiance-props}(a)) or its larger version (as in \cref{fig:scene-irradiance-props}(b))---the scene irradiance is identical in both cases.

Next, we consider how the scene irradiance varies with time. Suppose the sun illuminates the canyon as shown in \cref{fig:scene-irradiance-props}(c). We represent the scene as a collection of many infinitesimally small patches. Then, we can write the scene irradiance as the sum of the irradiances due to the patches in the sunlit area, the patches in the shadowed area, and the patches on the shadow boundary:
\begin{equation}
\begin{aligned}
    \Escene = 
    &\int \limits_{\mathclap{\wsunlit}} L(t; \vs) \left( \vn \cdot \vs \right) \dw \, + \\
    &\int \limits_{\mathclap{\wshadowed}} L(t; \vs) \left( \vn \cdot \vs \right) \dw \, + \\
    &\int \limits_{\mathclap{\wboundary}} L(t; \vs) \left( \vn \cdot \vs \right) \dw,
    \label{eq:Escene}
\end{aligned}
\end{equation}
where $L(t; \vs)$ is the time-varying radiance from the direction $\vs$, $\dw$ is the differential solid angle subtended by each scene patch, and $\wsunlit$, $\wshadowed$, and $\wboundary$ are the scene areas illustrated in \cref{fig:scene-irradiance-props}(c).

If we assume the scene to be Lambertian, then as the sun moves through the sky, the radiances from the patches in the sunlit and shadowed areas change very slowly.\footnote{This does not hold for mirror-like scene patches. The radiance $L(t; \vs)$ from a mirror-like patch changes dramatically when the patch reflects the sun toward the panel. However, since such specular reflections appear for short durations while we are interested in irradiance contributions over long durations, we will ignore their contributions in this analysis. } In addition, the sizes of the sunlit and shadowed areas ($\wsunlit$ and $\wshadowed$ in \cref{fig:scene-irradiance-props}(c)) change slowly as the sun moves across the canyon. Therefore, the first and second terms in \cref{eq:Escene} vary smoothly with time. The third term in \cref{eq:Escene} does not vary smoothly since the radiance values of the scene patches that lie on the boundary of the shadow change dramatically as the shadow passes over them. However, the shadow boundary is a linear geometric construct, and hence the size of the boundary area ($\wboundary$) is small. Specifically, if the scene is composed of $N^2$ patches, then the total area of the patches in the sunlit and shadowed areas is of order $N^2$, while the total area of the patches on the shadow boundary is of order $N$. Therefore, the contribution of the third term in \cref{eq:Escene} is negligible, and hence the scene irradiance $\Escene$ varies smoothly with time.

\subsection{Forecasting Irradiance Due to Scene Using a Single Image}
\begin{figure*}[t]
    \centering
    \includegraphics[width=\linewidth]{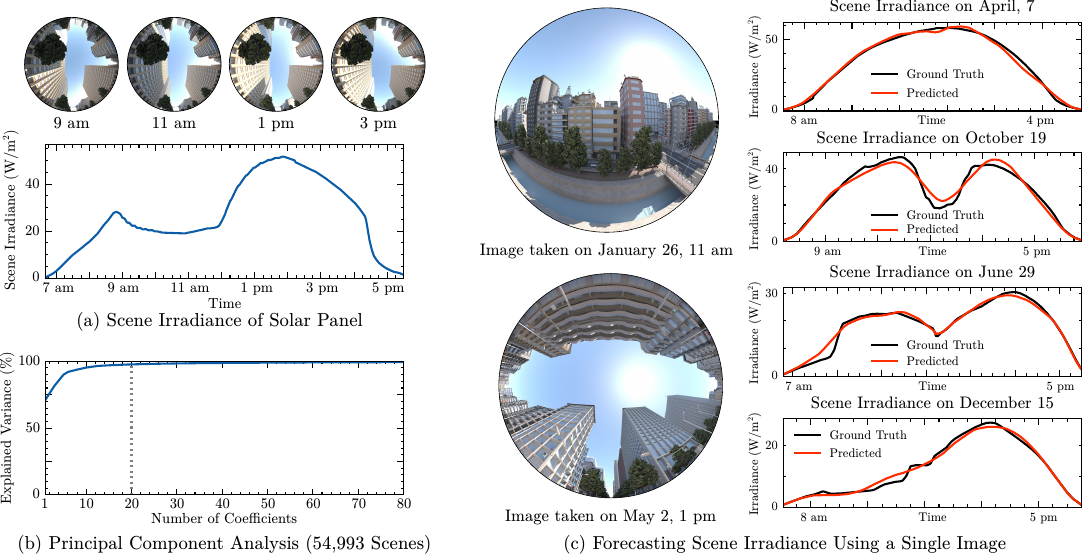}
    \caption{
        \textbf{Forecasting scene irradiance using a single image.}
        (a)~The scene irradiance of a solar panel in a simulated urban environment over the course of a day. Even though the illumination is complex, the panel irradiance due to the scene varies smoothly with time.
        (b)~We rendered the scene irradiance at 54,933 different locations for every sun position in the sky in a Lambertian urban environment and performed principal component analysis. The plot shows the explained variance in the scene irradiance as a function of the number of coefficients used to represent each function. From the plot, we see that 20 coefficients (dotted vertical line in the plot) are sufficient to capture $98\%$ of the variance in the scene irradiance for any sun position. This shows that the scene irradiance lies in a low-dimensional subspace, suggesting that a neural network could learn to predict the scene irradiance from a single image.
        (c)~We trained a neural network to predict the scene irradiance using a single hemispherical image. Prediction results from the neural network for two different rendered scenes are shown here. In each case, we predict the scene irradiance for two different days (with different sun paths). Note that the predicted scene irradiance plots (in red) closely follow the ground truth scene irradiance plots (in black) which were computed using a physically based renderer. 
    }
    \label{fig:scene-irradiance}
\end{figure*}

We used a physically based renderer~\cite{wenzeljakobMitsuba3Renderer2022} to simulate the scene irradiance in a Lambertian scene as a function of the sun's position at 54,933 urban locations on rooftops, on walls, and near the ground. \Cref{fig:scene-irradiance}(a) shows the simulated scene irradiance at one such location over the course of a day. Even though there are many objects casting shadows and producing reflections within the canyon, the scene irradiance in \cref{fig:scene-irradiance}(a) varies smoothly with time. 

We performed principal component analysis (PCA) on the 54,933 simulated scene irradiance functions and found that just 20 coefficients capture $98\%$ of the variance as a function of the sun's position (see the explained variance plot in \cref{fig:scene-irradiance}(b)). This indicates that the scene irradiance function lies in a low-dimensional subspace, which suggests that it could be forecasted from a single image.

Based on the above observations, we developed a neural network to predict the scene irradiance of a panel from a single image. The network contains a vision transformer backbone and an MLP head that outputs the scene irradiance for any sun position. The MLP consists of three hidden layers with sizes 512, 1024, and 2048 with ReLU activation functions. The input to the network is a single $256\times256 \, \unit{\text{px}\squared}$ HDR hemispherical image, along with the sky aperture and camera orientation $R_{EC}$ that are computed from it. Additionally, we provide the sun's position in the Earth coordinate frame at the time the image was captured as an additional input; this implicitly provides information about the brightness of the sun, which can be used to infer the reflectance of objects in the scene.
The sun's position in the sky is parameterized by its zenith and azimuth angles, which are uniformly discretized into $2.5\unit{\degree}$ bins. Using this discretization, the network outputs a $36\times144$ map, in which each element represents the scene irradiance for a given zenith and azimuth angle of the sun. We trained the network using 1.2 million rendered images in a simulated urban environment with clear sky conditions, along with the corresponding $36\times144$ scene irradiance functions.

\Cref{fig:scene-irradiance}(c) shows example outputs of the network for two rendered images. From a single hemispherical image, along with the sky aperture and camera orientation derived from it, the network learns to predict the scene irradiance for any sun position in the sky. In \Cref{fig:scene-irradiance}(c), we see that the predicted scene irradiance is remarkably close to the ground truth.

To quantify the accuracy of the network's prediction, we compare the error in the annual scene irradiance as a percentage of the total annual irradiance received by a panel. Across 276,828 images from 10,987 locations in the simulated urban environment, our approach forecasts the annual scene irradiance with an average error of $0.5\%$ of the total annual irradiance. If we instead ignored scene irradiance, we would underestimate the total annual irradiance by $12.3\%$. 
In \cref{app:scene-irradiance}, we provide additional details on the network training and report the network's performance when trained and tested in urban scenes with non-Lambertian reflectance models.

\begin{figure}[t]
    \centering
    \includegraphics[width=\linewidth]{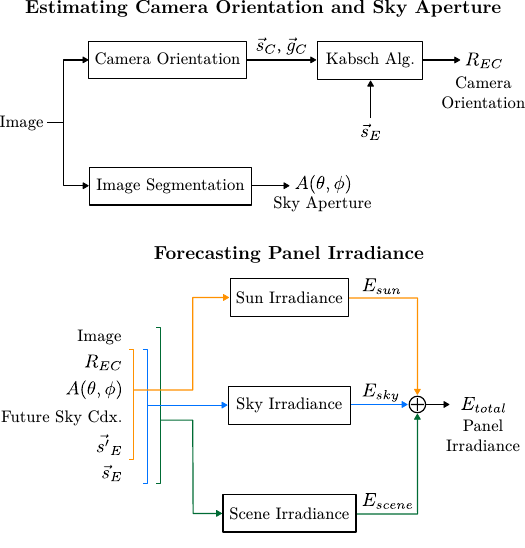}
    \caption{\textbf{The entire pipeline for forecasting a solar panel's irradiance using a single image.}
        First, we estimate the direction of the sun and gravity in the camera coordinate frame ($\vec{s}_C$ and $\vec{g}_C$, respectively) to determine the camera's orientation ($R_{EC}$). We use an image segmentation model to automatically determine the sky aperture ($A(\theta,\phi)$). To forecast the panel's irradiance at a future time, the sky conditions (parameterized by the GHI, DHI, and DNI) and sun position ($\vec{s'}_E$) at that time are provided as inputs. Using these inputs together with the image, the camera orientation, and the sky aperture, we forecast the panel's irradiance due to the sun ($E_{sun}$), the sky ($E_{sky}$), and the scene ($E_{scene}$). The total forecasted panel irradiance ($E_{total}$) is the sum of these three components. This entire process is automatic. The only requisite measurement is a single image, along with the time of day it was captured and the GPS location. 
    }
    \label{fig:flowchart}
\end{figure}

To summarize, our entire approach for forecasting irradiance from a single image is illustrated in \cref{fig:flowchart}. Using the single image, we first determine the direction of the sun and gravity in the camera coordinate frame ($\vec{s}_C$ and $\vec{g}_C$, respectively). The direction of the sun in the Earth coordinate frame ($\vec{s}_E$) is computed using the time the image was taken and its GPS location~\cite{michalskyAstronomicalAlmanacAlgorithm1988,meeusAstronomicalAlgorithms1991}. These corresponding vectors can be used to estimate the camera orientation $R_{EC}$. We use image segmentation to automatically determine the sky aperture $A(\theta,\phi)$. In order to forecast the panel's irradiance at any time in the future, the sky conditions (parameterized by the GHI, DHI, and DNI) at such time must be provided as inputs. Then, we use \cref{eq:Esun} to compute the irradiance due to the sun ($E_{sun}$) and \cref{eq:Esky} to compute the irradiance due to the sky ($E_{sky}$) at the future time. The scene irradiance ($E_{scene}$) is computed using the network described above. The total forecasted panel irradiance ($E_{total}$) is the sum of these three irradiance components. 

%% file: sections/sec05_experiments.tex
\begin{figure*}[p]
    \centering
    \includegraphics[width=\linewidth]{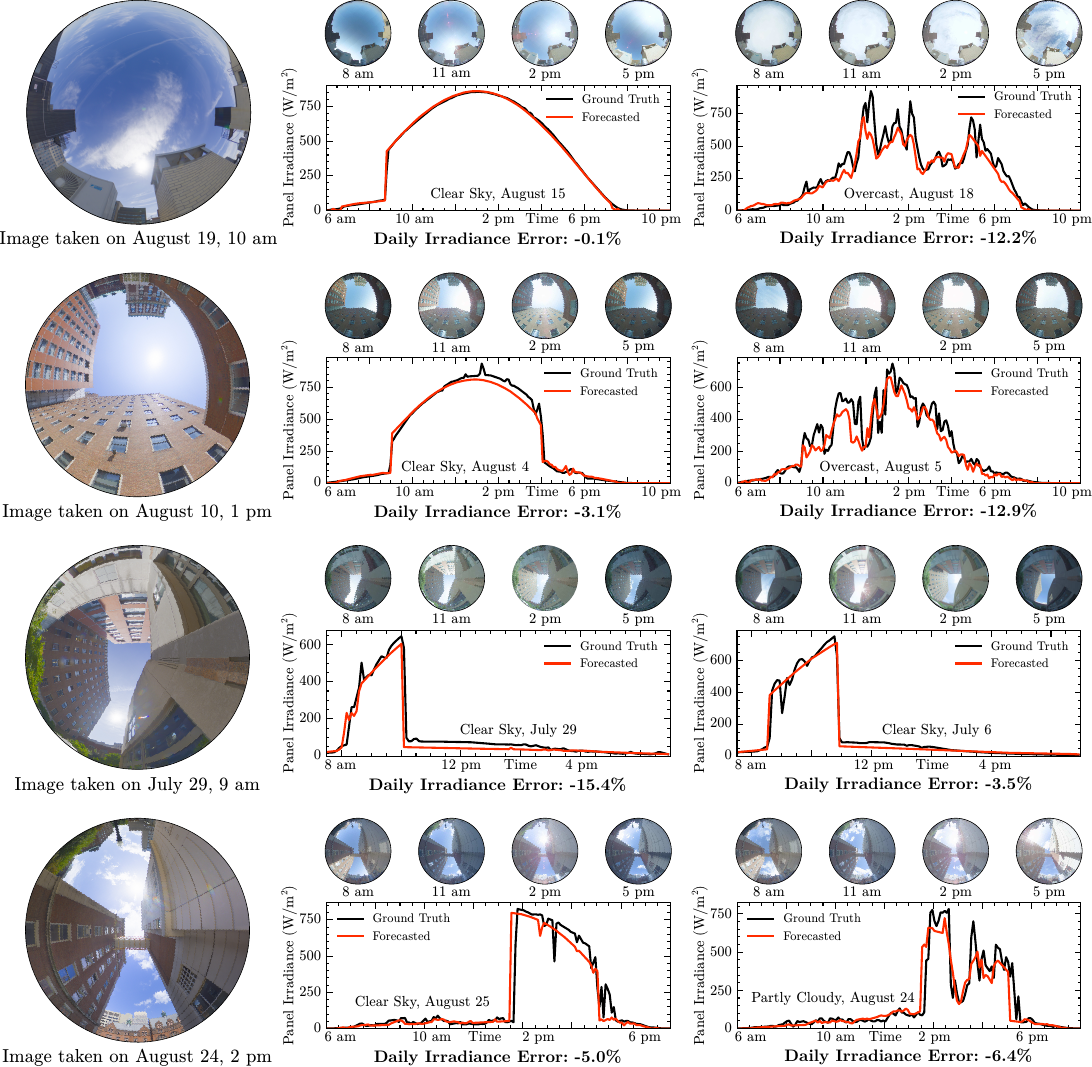}
    \caption{
        \textbf{Validating forecasted panel irradiance in urban settings.}
        We have tested our approach in four different locations: on a rooftop with an open view of the sky (first row), on a rooftop surrounded by tall buildings (second row), and in deep urban canyons where the contribution from the scene is significant (third and fourth rows). The single hemispherical image used to forecast the panel irradiance at each location is shown on the left. The plots on the right show the forecasted panel irradiance computed using our approach (red line) overlaid on the ground truth irradiance measured using a pyranometer (black line) for an entire day. The sky conditions for each day (clear sky, partly cloudy, and overcast sky) are listed within the plots. The fisheye images above the plots show snapshots of the illumination at different points in time. These inset images were only captured for visualization and were not used to forecast the panel irradiance. When the sky is clear, the step edges in the forecasted irradiance are close to the step edges in the ground truth irradiance, which correspond to the moments at which the sun moves in and out of the sky aperture. The forecasting error in the total daily irradiance is listed below each plot. 
    }
    \label{fig:experiments}
\end{figure*}

\section{Validating Forecasted Solar Panel Irradiance}
\label{sec:real-experiments}
We validated our approach in four different urban settings: on a rooftop with an open view of the sky (first row of \cref{fig:experiments}), on a rooftop surrounded by tall buildings (second row of \cref{fig:experiments}), and near the ground in deep urban canyons (third and fourth rows of \cref{fig:experiments}). At each location, we captured a single hemispherical image using a Ricoh Theta Z1 camera (first column of \cref{fig:experiments}) to forecast the panel irradiance, and we used a silicon pyranometer (EKO Instruments ML-01) to measure the ground truth irradiance.

In \cref{fig:experiments}, we plot the forecasted and ground truth irradiance at each location for various sky conditions. The inset images above the plots show the illumination seen by the panel. As can be seen from the inset images, when the sky is clear, the forecasted panel irradiance closely tracks the ground truth. Furthermore, our approach accurately predicts the step edges in the panel irradiance that correspond to the times at which the sun moves in and out of the sky aperture. This performance can be attributed, in part, to the accuracy of the estimated camera orientation.  In addition, notice that in deep urban canyons (third and fourth rows of \cref{fig:experiments}) the panel only sees the sun for a few hours in the day. In these cases, the sky and the scene contribute significantly to the total daily irradiance received by the panel.

The error in the forecasted irradiance received by the panel throughout each day is noted below each plot in \cref{fig:experiments}. In total, we have conducted experiments over 20 days at the four locations shown in \cref{fig:experiments}. The error in the forecasted daily irradiance on each day is tabulated in final column of \cref{tab:experiment-results}. Please refer to \cref{app:real-experiment-details} for experiment details and \cref{app:extended-results} for plots of the forecasted irradiance over longer time horizons (up to 9 days at a single location).

\section{Comparison with Alternative Approaches}
\label{sec:comparison}
We have compared the accuracy of our approach with an irradiance-based method for forecasting a panel's irradiance in the four urban environments shown in \cref{fig:experiments}. This alternative method computes the panel irradiance using a transposition model that decomposes the irradiance into sun (beam), sky (diffuse), and ground components. We have implemented this approach following the implementation of the System Advisor Model (SAM)~\cite{gilmanSAMPhotovoltaicModel2018}, a widely-used software tool for irradiance forecasting (see \cref{app:irradiance-based-implementation-details} for implementation details). The accuracy of this method's forecasted irradiance compared to ground truth is listed under ``Irradiance-Based Model 1'' in \cref{tab:experiment-results}. As can be seen in the table, this approach consistently underestimates the panel irradiance at all four locations. This is due, in part, to the fact that the sky (diffuse) irradiance is simply attenuated by the sky view factor to account for nearby buildings that obstruct the panel's view of the sky. This attenuation assumes an isotropic sky~\cite{gilmanSAMPhotovoltaicModel2018}. As a result, it does not account for the spatially-varying radiance from the sky (e.g.~circumsolar radiance), and hence it leads to large errors in urban canyons for which the sky contributes a significant portion to the total irradiance. Additionally, the ground irradiance component does not account for reflections from nearby structures~\cite{gilmanSAMPhotovoltaicModel2018}. Our approach, by contrast, uses a learned model to accurately forecast the irradiance due to the entire scene (reflections from both the ground and nearby buildings) using a single image.

In this first comparison, we have assumed that the sky aperture used to compute the irradiance components can be accurately obtained from an image captured by a camera with a known orientation. In a general urban setting, however, this is not the case and is precisely the problem our work seeks to solve. Instead, a common approach to determine the sky aperture is to render an image from the panel's location using a 3D model. The error in the forecasted irradiance using a transposition model along with a 3D model to determine the sky aperture is shown in \cref{tab:experiment-results} under the column ``Irradiance-Based Model 2.'' As we will show next, existing 3D models fail to capture the small structures near a panel that obstruct its view of the sky. This leads to an increase in the forecasted irradiance compared to using an accurate segmentation of the sky for determining the sky aperture (as in ``Irradiance-Based Model 1'' in \cref{tab:experiment-results}).

Prior works have also used an explicit 3D model of the environment to forecast the irradiance of a panel using ray tracing~\cite{jakubiecMethodPredictingCitywide2013, freitasModellingSolarPotential2015, mardaljevicIrradiationMappingComplex2003, zhuEffectUrbanMorphology2020, compagnonSolarDaylightAvailability2004, jakicaStateoftheartReviewSolar2018, kosmopoulosRayTracingModelingUrban2024, anSolarEnergyPotential2023, martinez-rubioEvaluatingSolarIrradiance2016, chengSolarEnergyPotential2020, andresTimevaryingRayTracing2023} and radiosity~\cite{robinsonSolarRadiationModelling2004}. This approach is implemented in widely-used software tools for assessing the solar energy potential of a surface~\cite{jakicaStateoftheartReviewSolar2018,googleProjectSunroof}. The accuracy of this approach, however, depends on the detail of the 3D model. We will show that existing 3D models fail to capture the tiny urban structures near a panel, causing the predictions to overestimate the panel's irradiance.

\begin{figure*}[t]
    \centering
    \includegraphics[width=\linewidth]{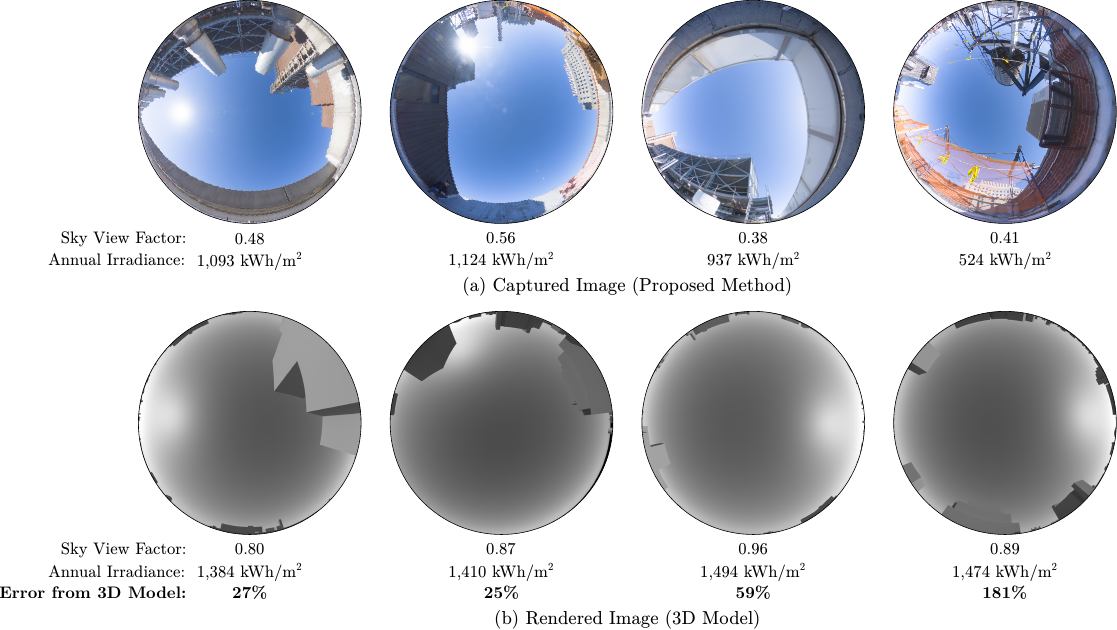}
    \caption{
        \textbf{Comparison with the irradiance forecasted using a 3D model.}
        (a)~A single image captured on an urban rooftop at four different locations. Small structures nearby such as parapets, vents, and HVAC units obstruct the view of the sky (see the sky view factor below each image). We used our approach to forecast the annual irradiance of a panel at each location, which is listed below each image.
        (b)~An image rendered using an existing 3D model of the environment at the four locations shown in (a). The 3D model does not capture all of the small structures near each panel, and hence the rendered images show an erroneously unobstructed view of the sky. This is reflected in the sky view factors of the rendered images, which overestimate the ground truth sky view factors computed using the captured images. We used ray tracing within the 3D model to forecast the panel's annual irradiance. 
        In these four examples, this approach for irradiance forecasting using a 3D model that fails to capture small structures leads to overestimates in the panel irradiance by up to $181\%$. 
    }
    \label{fig:3d_model}
\end{figure*}

Consider the left image in \cref{fig:3d_model}(a) taken on a city rooftop. There are multiple small structures next to the panel, including the roof's parapet and various vent stacks. Even though these structures are small, they are so close to the panel that they form a miniature canyon around it, thereby obstructing the panel's view of the sky. In this example, the structures in the image obstructed over half of the sky visible to the panel, yielding a sky view factor of $0.48$ (listed below in \cref{fig:3d_model}(a)). Given the significant impact these small structures have on the panel's view of the sky, it is critical to consider their effects when forecasting the panel irradiance. We used our approach to forecast the annual irradiance of the panel, which is listed below the image in \cref{fig:3d_model}(a).

The left image in \cref{fig:3d_model}(b) shows a rendered image from the same position and orientation as the left image in \cref{fig:3d_model}(a). Since the 3D model does not capture the smaller rooftop structures (e.g.,~the parapet, vent stacks, and HVAC unit), the rendered image incorrectly shows the panel having a significantly less obstructed view of the sky. In this example, the rendered image has a sky view factor of $0.80$ (listed below in \cref{fig:3d_model}(b)), significantly larger than the ground truth sky view factor of $0.48$ computed from the captured image. To demonstrate how this lack of detail in the 3D model translates to errors in the forecasted irradiance, we used a physically based renderer~\cite{wenzeljakobMitsuba3Renderer2022} to simulate the illumination seen by the panel via ray tracing. The annual irradiance computed using this approach is listed below the image in \cref{fig:3d_model}(b). 
Since the 3D model does not capture the small structures near the panel, this approach overestimates the panel's irradiance by $27\%$ compared to the irradiance computed using our approach. 

In \cref{fig:3d_model}, we compare four different images captured on urban rooftops with their corresponding images rendered using a 3D model. The model fails to capture the small structures near each panel, leading to overestimates of both the panel's sky view factor and its annual irradiance. Across the four images, the irradiance computed using the 3D model overestimates the irradiance predicted by our method by up to $181\%$. 

In addition, we have compared the accuracy of this approach with the ground truth irradiance from the real experiments in \cref{fig:experiments}. The error in the forecasted irradiance computed using ray tracing and a 3D model is tabulated in the ``3D Model'' column of \cref{tab:experiment-results}. Unlike existing 3D models, a single image captures all of the nearby structures that affect a panel's view of the sky and surrounding buildings, and hence our proposed approach often yields more accurate predictions. In \cref{app:extended-results-3d-model-comparison}, we provide details of the 3D model and compare the real and rendered images at the four locations.

In short, in urban settings where panels are almost guaranteed to be placed near small structures, the use of an existing 3D model to forecast the panel's irradiance yields inaccurate results. Since our approach uses an image taken from the panel's location, the irradiance forecasted by our method accounts for the effects of all the structures visible to the panel, regardless of their scale. In addition, the image captures a snapshot of the environment at the present time. This is not the case for existing 3D models that are typically updated every few years and hence do not capture the most recent changes to the urban landscape. Furthermore, our approach does not require ray tracing, which is computationally expensive. Instead, our approach relies on lightweight neural networks and illumination models that can be implemented on a mobile device, thus allowing a user to quickly forecast the irradiance of a panel once an image is taken. 

\begin{table*}[t]
    \centering
    \footnotesize
    \caption{
        \textbf{Error in the forecasted daily irradiance for the real experiments in \cref{fig:experiments}}. 
        For each day, we list the predominant sky conditions (clear sky, partly cloudy, or overcast). The error in the forecasted daily irradiance computed using our approach (Proposed) is shown in the final column. In addition, we have compared the performance of our approach with an irradiance-based approach and 3D model-based approach for irradiance forecasting. Irradiance-Based Model 1 uses a sky aperture computed from an image for which the camera orientation is determined using our approach. Irradiance-Based Model 2 compares the performance of the same method using a sky aperture rendered from a 3D model. The 3D model-based approach uses ray tracing to simulate the panel's illumination. The four sections in the table delineated by horizontal lines correspond to the four locations in \cref{fig:experiments}.
    }
    \vspace{0.1in}
    \label{tab:experiment-results}
    \begin{threeparttable}
        \begin{tabular}{@{}cccccc@{}}
            \toprule
            & & \multicolumn{1}{c}{} & \multicolumn{2}{c}{Error (\%)} & \multicolumn{1}{c}{} \\
            \cmidrule(lr){3-6}
            Day & Sky Conditions & \makecell{Irradiance-Based\\Model 1~\cite{gilmanSAMPhotovoltaicModel2018}\tnote{a}} & \makecell{Irradiance-Based\\Model 2~\cite{gilmanSAMPhotovoltaicModel2018}\tnote{b}} & 3D Model & Proposed \\
            \midrule
            July 6, 2025 & Clear & $-19.6$ & $-37.6$ & $-7.5$ & $-3.5$ \\
            July 29, 2025 & Clear & $-34.1$ & $-38.9$ & $-2.8$ & $-15.4$ \\
            \midrule
            August 2, 2025 & Partly Cloudy & $-17.5$ & $-12.4$ & $7.9$ & $0.5$ \\
            August 3, 2025 & Clear & $-12.0$ & $-7.7$ & $6.4$ & $0.0$ \\
            August 4, 2025 & Clear & $-20.1$ & $-15.3$ & $3.4$ & $-3.1$ \\
            August 5, 2025 & Overcast & $-46.9$ & $-41.9$ & $-5.9$ & $-12.9$ \\
            August 6, 2025 & Overcast & $-59.8$ & $-54.9$ & $-20.2$ & $-27.0$ \\
            August 7, 2025 & Partly Cloudy & $-21.2$ & $-14.4$ & $8.0$ & $-1.0$ \\
            August 8, 2025 & Clear & $-9.4$ & $-2.9$ & $9.6$ & $0.5$ \\
            August 9, 2025 & Clear & $-9.5$ & $-2.9$ & $9.9$ & $-0.1$ \\
            August 10, 2025 & Clear & $-11.9$ & $-5.7$ & $8.5$ & $-0.3$ \\
            \midrule
            August 15, 2025 & Clear & $-4.3$ & $5.1$ & $6.6$ & $-0.1$ \\
            August 16, 2025 & Partly Cloudy & $-9.6$ & $1.1$ & $4.8$ & $-0.9$ \\
            August 17, 2025 & Partly Cloudy & $-10.8$ & $2.1$ & $5.2$ & $-3.6$ \\
            August 18, 2025 & Overcast & $-27.8$ & $-10.9$ & $-4.8$ & $-12.2$ \\
            \midrule
            August 24, 2025 & Partly Cloudy & $-20.8$ & $-22.0$ & $1.2$ & $-6.4$ \\
            August 25, 2025 & Clear & $-12.9$ & $-15.1$ & $0.9$ & $-5.0$ \\
            August 26, 2025 & Partly Cloudy & $-20.1$ & $-20.6$ & $5.7$ & $-3.0$ \\
            August 27, 2025 & Partly Cloudy & $-18.7$ & $-22.8$ & $2.0$ & $0.2$ \\
            August 28, 2025 & Partly Cloudy & $-34.4$ & $-45.2$ & $-9.8$ & $-10.3$ \\
            \bottomrule
        \end{tabular}

        \begin{tablenotes}
            \footnotesize
            \item[a] The sky aperture and sky view factor are computed from the single image captured by our approach.
            \item[b] The sky aperture and sky view factor are computed by rendering an image using a 3D model.
        \end{tablenotes}
    \end{threeparttable}
\end{table*}

%% file: sections/sec06_citibike.tex
\section{Finding the Best Orientation of a Solar Panel}
A common rule of thumb states that the best fixed orientation of a panel is pointing toward the equator with a zenith angle equal to the panel's latitude~\cite{duffieSolarEngineeringThermal2006}. This assumes the panel has a clear view of the sky, which is rarely the case in urban settings. Instead, we show that a single spherical image taken near an existing panel can be used to identify the best fixed panel orientation, which may differ substantially from the above rule of thumb.

\subsection{Determining Panel Orientation in the Camera Coordinate Frame}
\begin{figure*}[t]
    \centering
    \includegraphics[width=\linewidth]{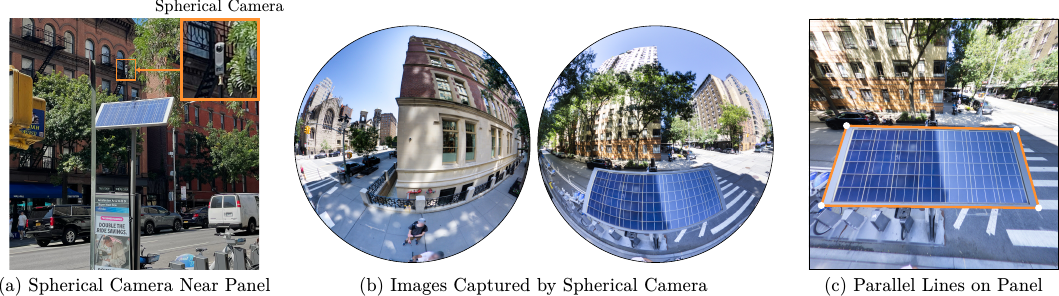}
    \caption{
        \textbf{Determining the orientation of a panel from a spherical image.}
        (a)~A spherical camera placed directly above an existing panel. 
        (b)~The camera simultaneously captures two hemispherical images with opposing views. Notice that the panel is visible in one of the two images captured by the camera. 
        (c)~A perspective view of the panel generated from the hemispherical images. We use the locations of the panel's corners (white circles) to determine the lines along the panel's boundary (orange lines). The vanishing points of the two pairs of boundary lines reveal the panel's orientation in the camera coordinate frame.
    }
    \label{fig:spherical}
\end{figure*}

Consider the panel powering a bike share station in \cref{fig:spherical}(a). We placed a Ricoh Theta Z1 camera directly above the panel to simultaneously capture two hemispherical images from opposing views (\cref{fig:spherical}(b)).  As can be seen in \cref{fig:spherical}(b), the panel appears in one of the hemispherical images captured by the camera. It turns out that we can use the lines along the panel boundary to determine the orientation of the panel in the camera coordinate frame. To illustrate this, \cref{fig:spherical}(c) shows a perspective image of the panel in which we have manually identified the locations of the panel corners (white circles in \cref{fig:spherical}(c)). From the corners, we can determine the lines along the panel's boundary (orange lines in \cref{fig:spherical}(c)). If we assume the panel to be rectangular, then each pair of opposite lines on the boundary corresponds to parallel lines in the scene. Therefore, we can use the vanishing points of the line pairs to determine their orientation in the camera coordinate frame~\cite{hartleyMultipleViewGeometry2003}. Once we know the orientation of the line pairs, we can compute the panel's normal vector up to a sign ambiguity, as it is orthogonal to the two pairs of lines. We resolve the sign ambiguity by enforcing that the panel be oriented above the horizon, which is almost certainly the case in any real world installation. Please refer to \cref{app:panel-orientation} for further details of this method.

\subsection{Finding the Best Orientation of Existing Panels}
\begin{figure*}[p]
    \centering
    \includegraphics[width=\linewidth]{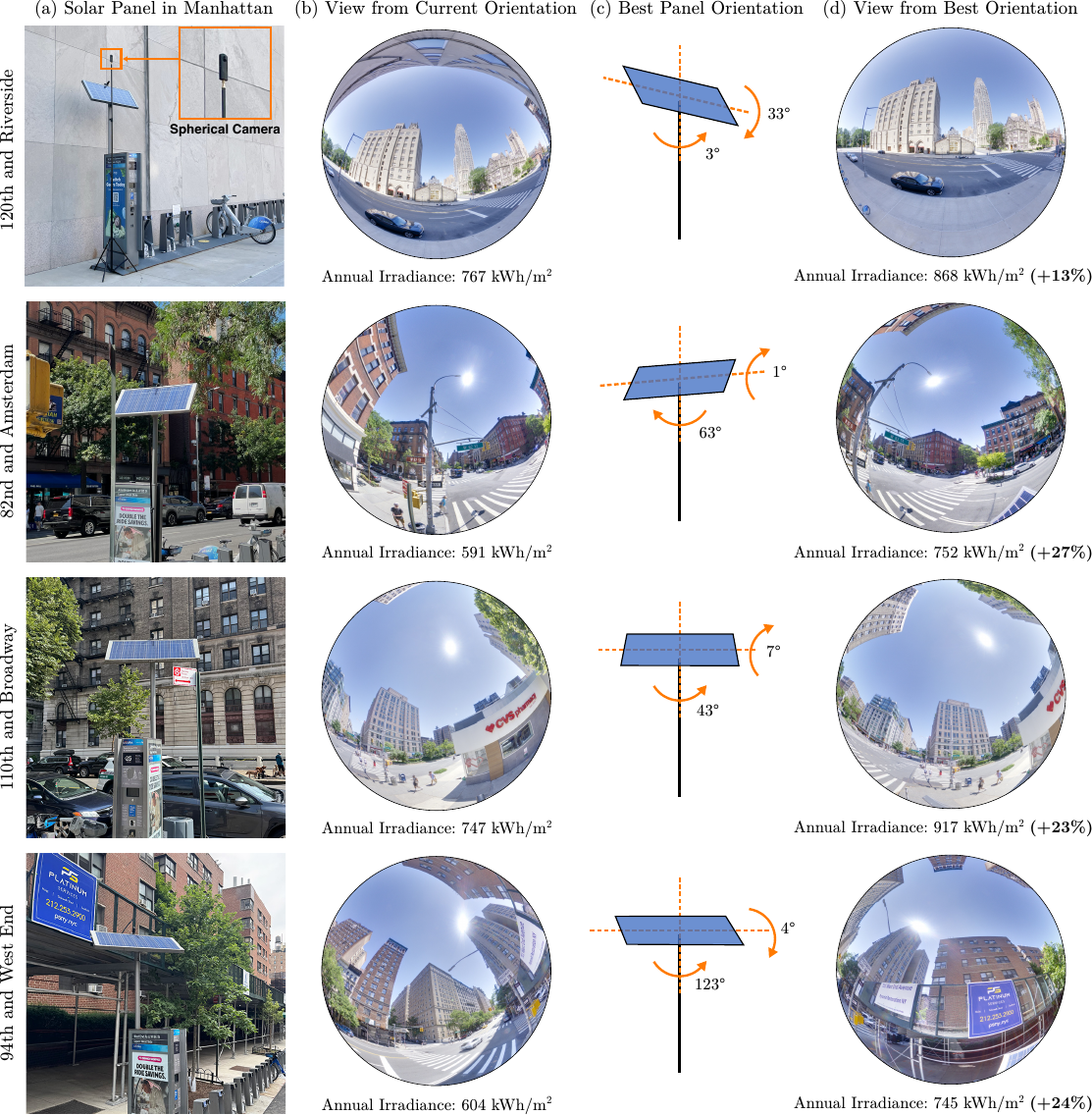}
    \caption{
        \textbf{Finding the best orientation of existing panels in Manhattan.}
        (a)~We placed a spherical camera directly above four different panels in use in Manhattan and captured a single spherical image. 
        (b)~Since the panel is visible to the camera, we can determine the orientation of the panel in the camera coordinate frame. This allows us to generate the hemispherical image seen by the panel in its current orientation. Using this image, we applied our approach to estimate the total annual irradiance received by the panel. Next, we searched for the best panel orientation by generating a hemispherical image for all possible orientations and applying our approach to each image to forecast the irradiance.
        (c)~The adjustments needed to orient each panel in the direction that yields maximum annual irradiance. 
        (d)~The view from each panel for the optimal orientation and the corresponding forecasted annual irradiance. None of the four panels were installed close to the optimal orientation---we estimate that rotating the panels by the angles in~(c) would increase the annual irradiance by up to $27\%$.
    }
    \label{fig:citibike}
\end{figure*}

We have taken a spherical image near four existing solar panels in Manhattan (\cref{fig:citibike}(a)). By using the lines along each panel's boundary to determine its orientation in the camera coordinate frame, we can generate the hemispherical image seen by each panel in its current orientation (\cref{fig:citibike}(b)). Then, we apply our approach for forecasting irradiance to the hemispherical image to estimate the total annual irradiance received by each panel in a typical year. These estimates are noted below each image in \cref{fig:citibike}(b). 

Since we captured a spherical image, we can also generate a hemispherical image for any panel orientation. We can therefore search for the hemispherical image (i.e.,~panel orientation) that yields the maximum annual irradiance. \Cref{fig:citibike}(c) shows the adjustments (rotations) to the panel orientation that would be needed to arrive at the optimal panel orientation, and \cref{fig:citibike}(d) shows the corresponding hemispherical images. Below each image in \cref{fig:citibike}(d) we have noted the total annual irradiance for the optimal orientation and the percentage gain in irradiance (and hence harvested energy) that would result from using the optimal orientation. These results reveal that none of the four panels shown in \cref{fig:citibike}(a) are oriented in the optimal direction. All the panels need to be rotated by significant angles, and we estimate that these adjustments would increase the annual irradiance by up to $27\%$.

%% file: sections/sec07_solaris.tex
\section{\textit{Solaris}: A Capture Device for Forecasting Panel Irradiance}
\begin{figure*}[t]
    \centering
    \includegraphics[width=\linewidth]{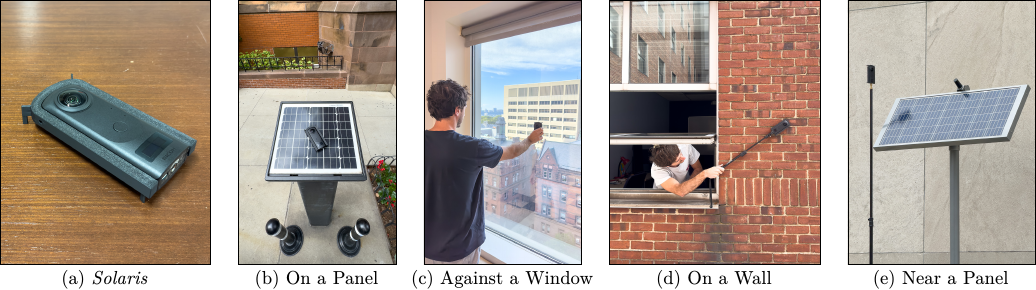}
    \caption{
        \textbf{\textit{Solaris:} a device for capturing images for irradiance forecasting.} 
        (a)~\textit{Solaris} combines an existing camera with a chassis that makes it easy to capture the hemispherical and spherical images our approach uses for irradiance forecasting. Four pegs on the corners of the device keep it parallel to any flat surface it is placed on or against. 
        \textit{Solaris} can be (b)~placed on a horizontal panel or surface, (c)~held against a window, or~(d) placed against a building wall to capture hemispherical images. (e)~It can also be placed near an existing panel to capture a spherical image.
    }
    \label{fig:device}
\end{figure*}

We developed \textit{Solaris}, a device for easily capturing images that can be used to forecast the irradiance of a variety of urban surfaces. \textit{Solaris} combines an existing spherical camera with a chassis that includes four corner pegs to keep the device parallel to flat surfaces, as shown in \cref{fig:device}(a). \Cref{fig:device}(b)-(e) illustrates four different use cases: \textit{Solaris} can be placed on a horizontal panel or surface (\cref{fig:device}(b)), held against a window (\cref{fig:device}(c)), held against a wall (\cref{fig:device}(d)), or placed near an existing panel (\cref{fig:device}(e)). In the future, a similar device could be implemented on a smartphone by using lens adapters to convert the phone's front- and back-facing cameras into hemispherical cameras.

%% file: sections/sec08_discussion.tex
\section{Discussion}
We have presented an approach for forecasting the irradiance of a solar panel using a single hemispherical image. Before a panel is installed, our approach can be used to quickly assess the solar energy potential of an urban surface, thereby reducing the soft costs associated with the panel installation. Once a panel is installed, our approach can forecast the temporal variation of the panel's irradiance, which is valuable information for the electric grid. We also showed that a spherical image can be used to find the panel orientation that yields maximum irradiance.
There are two directions we plan to pursue in the future. First, we plan to work with teams leading infrastructure projects in New York who have expressed an interest in using our approach, at-scale, to design solar energy systems for new constructions. On the research front, we are interested in exploring the idea of forecasting irradiance using images taken from a distance from the panel location of interest. This is a hard problem that requires estimating the sky aperture and the properties of an urban canyon corresponding to a location without making measurements at the location. Once again, we believe that a variety of advanced methods in computer vision may help us achieve this goal.  

\section{Limitations}
Our approach assumes that the image is captured on a sunny day. Although the sun does not need to be visible in the image, the method relies on visual cues from shadows to determine the direction of the sun and, consequently, the orientation of the camera. As a result, our method cannot be applied to images captured on overcast days when neither the sun nor shadows are visible. Our approach also uses a sky model to estimate the illumination seen by the panel. While this model accurately represents illumination under spatially smooth sky conditions (e.g.,~clear, overcast, or hazy skies), it cannot capture the effects of small clouds that briefly obstruct the sun. Additionally, this does not model nighttime illumination from city infrastructure. Our method further assumes that the scene will not change once the image has been captured. In practice, this is not always the case. For example, panels installed near trees may experience seasonal changes in illumination as leaves blossom and fall. Similarly, nearby vehicles or new construction may alter the scene geometry and hence affect a panel's irradiance. Since our method relies on a single image captured at one moment in time, it does not account for such changes. Finally, our experimental validation relies on estimates of sky conditions derived from satellite imagery to forecast irradiance. These estimates are known to be inaccurate and therefore introduce errors in the forecasted irradiance that are unrelated to the proposed approach.

%% file: sections/acknowledgements.tex
\section*{Acknowledgements}
This work was supported by the Office of Naval Research (ONR) under award number N00014-23-1-2096. Jeremy Klotz was supported by a National Defense Science and Engineering Graduate (NDSEG) Fellowship. The authors are grateful to Behzad Kamgar-Parsi at ONR for supporting this work and Vijay Modi at Columbia University for valuable technical feedback. The authors also thank Vlad Pyltsov for helpful discussions.

%% file: sections/app01_pose.tex
\section{Details of Estimating Camera Orientation}
\label{app:pose}

\subsection{Rendered Dataset of Hemispherical Images in Urban Settings}
\label{app:rendered-dataset}
We created a large scale dataset of rendered hemispherical images in an urban environment to train the neural networks for estimating the camera orientation and the scene irradiance. The images were rendered using Mitsuba~\cite{wenzeljakobMitsuba3Renderer2022}, a physically based renderer, and a 3D model of a city. The dataset consists of hemispherical images taken at 54,933 typical solar panel locations  in the simulated urban environment (near the ground, on walls, and on rooftops). At each location, we rendered multiple images for different positions of the sun in the sky. In total, the dataset consists of 1,382,400 hemispherical images.

To model the illumination from the sky, we used the Hosek-Wilkie sky model~\cite{hosekAnalyticModelFull2012} with a turbidity of $3$. This model allows us to both render the RGB image that would be captured by a camera and compute the total irradiance received by the panel over the entire spectrum. We model the sun as a distant point source whose brightness is given by the direct normal irradiance (DNI) on a clear day~\cite{ineichenNewAirmassIndependent2002,perezNewOperationalModel2002}. All objects in the scene are assumed to be Lambertian, and the images are rendered in Mitsuba using a path tracer.

\subsection{Network Training}
The input to the network is a $512\times512\,\unit{\text{px}}$ HDR image. Given that the dynamic range of outdoor images can be extremely large, we apply the pixel-wise transform $\log(x + 1)$ to the linear HDR image to compress its dynamic range before passing it into the network. We pre-trained the network for estimating the camera orientation using the rendered dataset described above. We used the cross entropy loss for each of the four predicted angles, and the network was trained using the AdamW optimizer~\cite{loshchilovDecoupledWeightDecay2019} with a learning rate of $5\cdot10^{-5}$.

We fine-tuned the network on real images taken in urban environments from the \textit{UrbanSky} dataset~\cite{klotzMinimalSensingOrienting2025}. To generate hemispherical images from \textit{UrbanSky} with known sun and gravity vectors, we manually identified 273 panoramas in the dataset for which the sun is visible to the camera. We then split the 273 panoramas from \textit{UrbanSky} into a set of 191 panoramas for training, 31 for validation, and 51 for testing. For each panorama, we randomly generated hemispherical images with many orientations. The ground truth sun vector is computed using the location of the sun in the original panorama, even if the sun is not visible in the hemispherical image. The ground truth gravity vector is computed from the orientation of each hemispherical image, since the gravity vector in each \textit{UrbanSky} panorama is known.

In total, this process creates 93,622 hemispherical images for training, 15,500 images for validation, and 25,415 images for testing. We fine-tuned the network on these images from \textit{UrbanSky} using the AdamW optimizer~\cite{loshchilovDecoupledWeightDecay2019} with a learning rate of $5\cdot10^{-5}$.

\subsection{Estimating Camera Orientation Using Gravity Vector from IMU}
If the camera contains an IMU, we can assume that the accelerometer accurately measures the direction of gravity in the camera coordinate frame. In this case, we modified the existing network to take the direction of gravity as an additional input rather than predicting the gravity vector using an MLP. To this end, we concatenated the zenith and azimuth angles of the gravity vector to input of the MLP that predicts the sun's direction. We then fine-tuned the entire network first with the rendered data (with a learning rate of $10^{-5}$) and then with the \textit{UrbanSky} dataset (with a learning rate of $10^{-4}$).

\subsection{Computing $R_{EC}$ When Gravity is Measured by the IMU}
If the gravity vector in the camera coordinate frame is known with high accuracy (i.e.,~it is measured by an IMU), we use an alternative method to compute $R_{EC}$ to ensure that the rotation exactly aligns the two gravity vectors. Let $R_{EC}'$ be a rotation matrix that rotates the gravity vector in the camera coordinate frame to the corresponding vector in the Earth coordinate frame.
Naturally, the rotation $R_{EC}'$ is not uniquely determined by a single pair of vectors. The rotation still has an unknown degree of freedom that corresponds to the azimuth angle of the camera $\phi$ in the Earth coordinate frame. We use the sun vectors in the two coordinate frames to compute $\phi$ by minimizing the objective:
\begin{equation}
    \phi = \arg\min_{\phi'}  \ \lVert u(\phi') \, R_{EC}' \, \vec{s_c} - \vec{s_e} \rVert^2,
\end{equation}
where $u(\phi')$ is a rotation about the $z$-axis by $\phi'$, $\vec{s_c}$ is the sun vector in the camera coordinate frame, and $\vec{s_e}$ is the sun vector in the Earth coordinate frame. The final rotation is then computed as:
\begin{equation}
    R_{EC} = u(\phi) \, R_{EC}'.
\end{equation}

\subsection{Estimating Camera Orientation When the Sun is Not Visible}
\begin{figure*}[p]
    \centering
    \includegraphics[width=\linewidth]{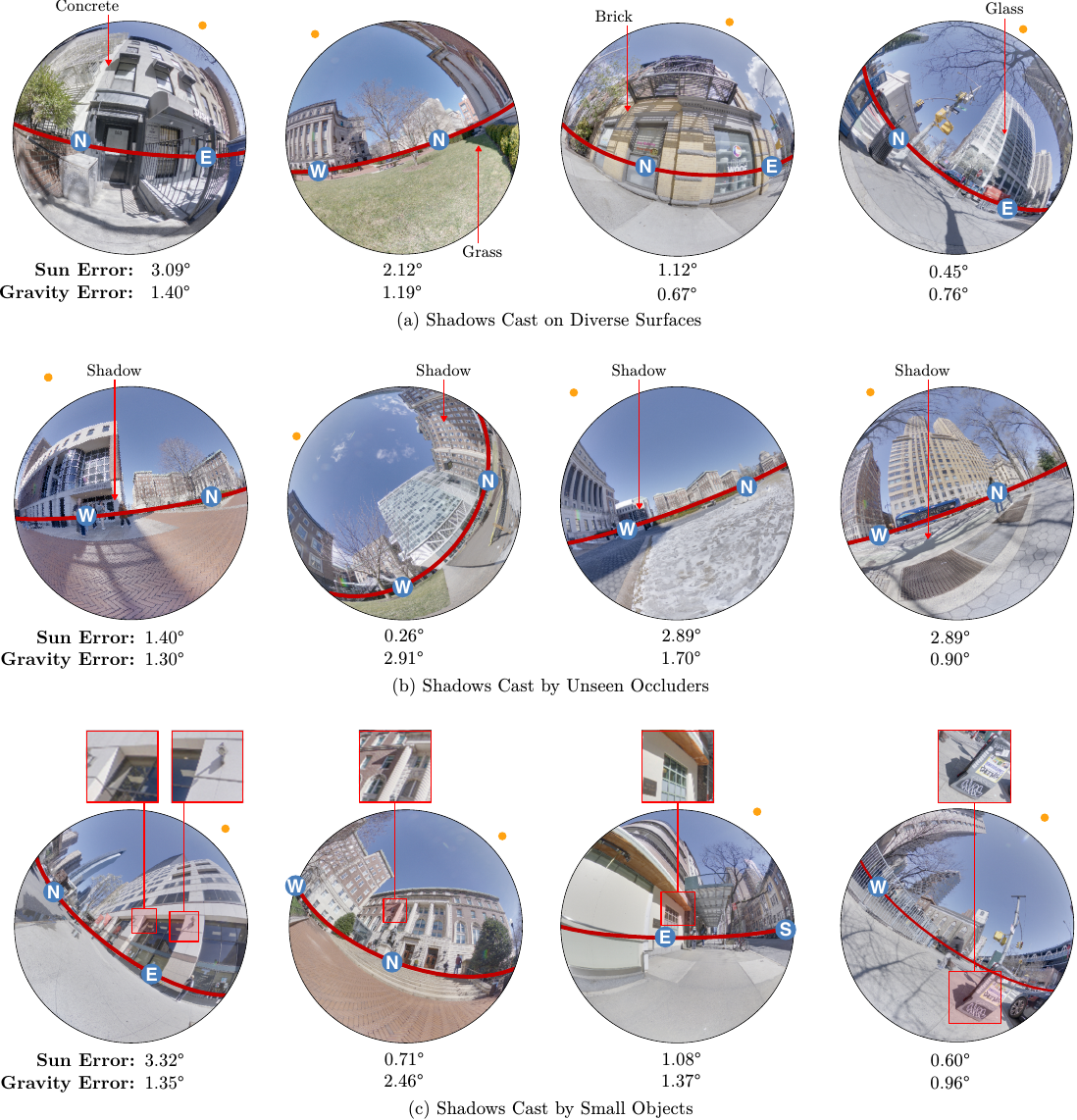}
    \caption{
        \textbf{Estimating the camera orientation when the sun is not visible in the image.}
        (a)~Results in scenes with shadows cast on diverse surfaces (e.g.,~concrete, grass, bricks, glass, and uneven surfaces). The orange dot denotes the predicted direction of the sun, and the red line denotes the predicted horizontal line (orthogonal to the direction of gravity). The accuracy of the predicted vectors is listed below each image. The network accurately predicts the direction of the sun and gravity even when shadows are cast on complex textures and uneven surfaces.
        (b)~Results in scenes with a large shadow cast by an object (occluder) that is not completely visible in the image. These shadows still provide useful cues to determine the sun's direction since they were likely cast by vertical or horizontal scene edges.
        (c)~Hemispherical images taken in urban environments also contain a plethora of small shadows cast by objects such as poles, lights, and window sills. Shown here are images that only contain small shadows (see inset images), without any large shadows cast by nearby buildings. These small shadows serve as additional visual cues for identifying the direction of the sun.
    }
    \label{fig:supp-pose}
\end{figure*}

\Cref{fig:supp-pose} shows the output of the network for scenes in which the sun is not visible in the image. As can be seen in \cref{fig:supp-pose}(a), the network accurately identifies the direction of the sun and gravity in the presence of shadows cast on complex textures and uneven surfaces. This performance can be attributed, in part, to the fact that the network has been trained using hemispherical images derived from 191 different scenes in \textit{UrbanSky}. Since these scenes were captured at various locations in New York City and at different times of day, the training data includes surfaces with diverse appearance and structure. As a result, the network is robust to such variations.

In some cases, the occluder that cast a shadow may not be visible in the image. Examples of this scenario are shown in \cref{fig:supp-pose}(b). In all four examples, there is a single large shadow cast by a nearby object (occluder) that is not fully visible. However, the shape of the shadow still provides a useful cue for determining the sun's direction since the shadow was likely cast by a vertical or horizontal edge on a nearby urban structure.

Additionally, any image taken in an urban environment likely contains multiple shadows that are cast by small objects. \Cref{fig:supp-pose}(c) shows four such examples. In each scene, small objects such as lights, poles, window sills, and protrusions from building walls cast tiny shadows. These small shadows together provide an additional cue for accurately determining the sun's direction when the sun is not visible in the image.

\subsection{Network Performance as a Function of Sky View Factor}
\begin{figure}[t]
    \centering
    \includegraphics[width=0.8\linewidth]{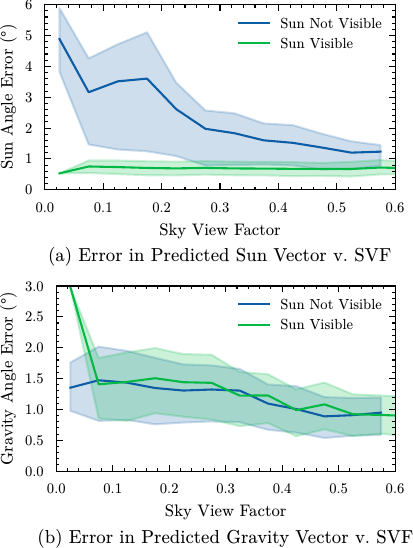}
    \caption{
        \textbf{Error in the predicted sun and gravity vectors as a function of the sky view factor (SVF).}
        (a)~When the sun is visible (green line), the network predicts the direction of the sun with high accuracy (less than 1 degree, on average). When the sun is not visible (blue line), the error in the predicted sun vector increases as the SVF decreases. This suggests that even when the sun is not visible to the camera, the brightness pattern of the sky provides an additional cue for determining the sun's direction. 
        (b)~The predicted gravity vector is less sensitive to the SVF, since the direction of gravity can be determined using horizontal and vertical scene lines on urban structures. The shaded region around each line denotes the interquartile range. 
    }
    \label{fig:supp-pose-svf}
\end{figure}

\Cref{fig:supp-pose-svf} shows the average error in the predicted sun and gravity vectors as a function of the sky view factor (SVF) across all 25,415 test images. When the sun is visible in the image (green line), the network predicts the direction of the sun with an average error of less than a degree, regardless of the sky view factor. When the sun is not visible, the average error increases with decreasing SVF. These results suggest that when the sun is not visible, the network is using visual cues from both the shadows on nearby structures and the brightness pattern of the sky to determine the sun's direction. \Cref{fig:supp-pose-svf}(b) shows the average error in the estimated gravity vector as a function of the SVF. The performance is less sensitive to the SVF since vertical and horizontal lines on nearby structures provide a strong cue for determining the direction of gravity. 

%% file: sections/app07_sky_model.tex
\section{Details of the Sky Model}
\label{app:sky-model}
We use the Perez sky model~\cite{perezAllweatherModelSky1993} in \cref{eq:Esky} to compute the radiance distribution from the sky at any future time. The sky conditions are parameterized by the DNI and DHI. From these irradiance measurements, we compute the model coefficients $a$, $b$, $c$, $d$, and $e$ as described by Perez \etal~\cite{perezAllweatherModelSky1993}. These coefficients are a function of the brightness and clearness indices, and they describe the overall shape of the sky's radiance distribution.

Following Perez \etal~\cite{perezAllweatherModelSky1993}, the radiance distribution of the sky is computed up to a scale factor as
\begin{equation}
    \ell_{sky}(\theta,\phi) = \left(1 + a e^{b / \cos(\theta)}\right) \left( 1 + ce^{d\gamma} + e\cos^2\gamma \right),
\end{equation}
where $\theta$ and $\phi$ are the zenith and azimuth angles of the current sky element, and $\gamma$ is the angle between the current sky element and the sun. 
The final radiance distribution $L(\theta,\phi)$, in $\unit{\watt\per\steradian\per\meter\squared}$, is computed by normalizing $\ell_{sky}$ such that the irradiance due to the upper hemisphere is equal to the DHI,
\begin{equation}
    L_{sky}(\theta,\phi) = \frac{E_{DHI}}{\int\limits_{0}^{2\pi} \int\limits_{0}^{\frac{\pi}{2}} \ell_{sky}(\theta', \phi') \, \cos\theta' \, d\theta' \, d\phi'} \, \ell_{sky} (\theta, \phi).
\end{equation}

%% file: sections/app02_scene_irradiance.tex
\section{Details of Forecasting Scene Irradiance}
\label{app:scene-irradiance}

\subsection{PCA of Scene Irradiance Functions}
The sun's position in the sky is parameterized by its zenith and azimuth angles, where each angle is uniformly discretized into $2.5\unit{\degree}$ bins. Using this discretization, the scene irradiance as a function of the sun position is represented by a $36\times144$ map. We performed PCA on this $36\times144$ representation of each scene irradiance function and found that just 20 coefficients capture $98\%$ of the variance in the scene irradiance.

\subsection{Inputs and Outputs of the Network}
The input to the network for estimating scene irradiance is a four channel image stack, where the first three channels correspond to the high dynamic range $256\times256 \, \unit{\text{px}\squared}$ color image, and the last channel corresponds to the $256\times256 \, \unit{\text{px}\squared}$ sky aperture. The camera orientation $R_{EC}$ and the position of the sun in the sky (a unit vector in the Earth coordinate frame) is concatenated with the image features from the backbone to form the input of the MLP. The output of the model is a 5184-length vector, which is reshaped into the $36\times144$ map representing the scene irradiance as a function of the sun's position.

\subsection{Network Training}
We trained the network to estimate the scene irradiance using the rendered dataset described in \cref{app:rendered-dataset}. Alongside each rendered hemispherical image, we have also computed the sky aperture, camera orientation $R_{EC}$, and the ground truth scene irradiance as a function of the sun position (represented as a $36\times144$ map).

We split the dataset of rendered images into 967,520 hemispherical images for training, 138,052 images for validation, and 276,828 images for testing. The network was trained using the Huber loss, the AdamW optimizer, and a learning rate of $10^{-5}$.


\subsection{Evaluation With Non-Lambertian Scenes}
We have shown that for a Lambertian scene, the scene irradiance of a panel varies slowly over time. However, real objects are non-Lambertian and produce complex illumination effects such as specular reflections, roughness, and gloss. Here we evaluate the performance of the model for predicting scene irradiance in non-Lambertian scenes. We rendered the dataset of hemispherical images at $54,933$ urban locations using non-Lambertian materials to train and test the network. The reflectance model used for each object is specified in the original 3D model, which is available online~\cite{JapaneseCityMidtown}. 

We used this rendered dataset to train and test the network for estimating a panel's scene irradiance from a single image. Using the network to predict the scene irradiance yields an average error in the total annual irradiance of $0.6\%$. By comparison, if we ignored scene irradiance, we would underestimate the annual panel irradiance by $14.5\%$. This illustrates that the network can estimate the scene irradiance in the presence of complex surface appearances.

%% file: sections/app04_experiment_details.tex
\section{Details of Forecasted Irradiance Validation}
\label{app:real-experiment-details}
\subsection{Sky Conditions Used in Real Experiments}
Our approach forecasts the panel irradiance using sky conditions parameterized by the GHI, DHI, and DNI. For the days when the ground truth irradiance was measured, we obtained satellite-derived estimates of these quantities from Solcast. Given that these satellite-derived estimates are known to be noisy~\cite{brightSolcastValidationSatellitederived2019,yangWorldwideValidation82020}, any errors in these three irradiances (GHI, DHI, and DNI) manifest as errors in the forecasted panel irradiance. This leads to errors in the forecasted panel irradiance that are not due to a limitation of our approach. For example, in \cref{tab:experiment-results}, the error in the forecasted panel irradiance from all of the approaches is higher on overcast days. This suggests that the errors in the forecasted irradiance on these days may be due to errors in the estimated sky conditions rather than a limitation of all three approaches.

\subsection{Forecasting Scene Irradiance in Real Experiments}
Since the model for estimating the scene irradiance was trained on scenes with a clear sky, we approximate the scene irradiance in other sky conditions by scaling the predicted scene irradiance by
\begin{equation}
    R = \frac{GHI_{current}}{GHI_{clear}},
\end{equation}
where $GHI_{current}$ is the global horizontal irradiance (GHI) in the current sky conditions, and $GHI_{clear}$ is the GHI in the clear sky generated by the Hosek-Wilkie model. 

%% file: sections/app03_extended_results.tex
\section{Experiments in Real Urban Canyons}

\subsection{Results over Longer Time Horizons}
\label{app:extended-results}

\begin{figure*}[p]
    \centering
    \includegraphics[width=\linewidth]{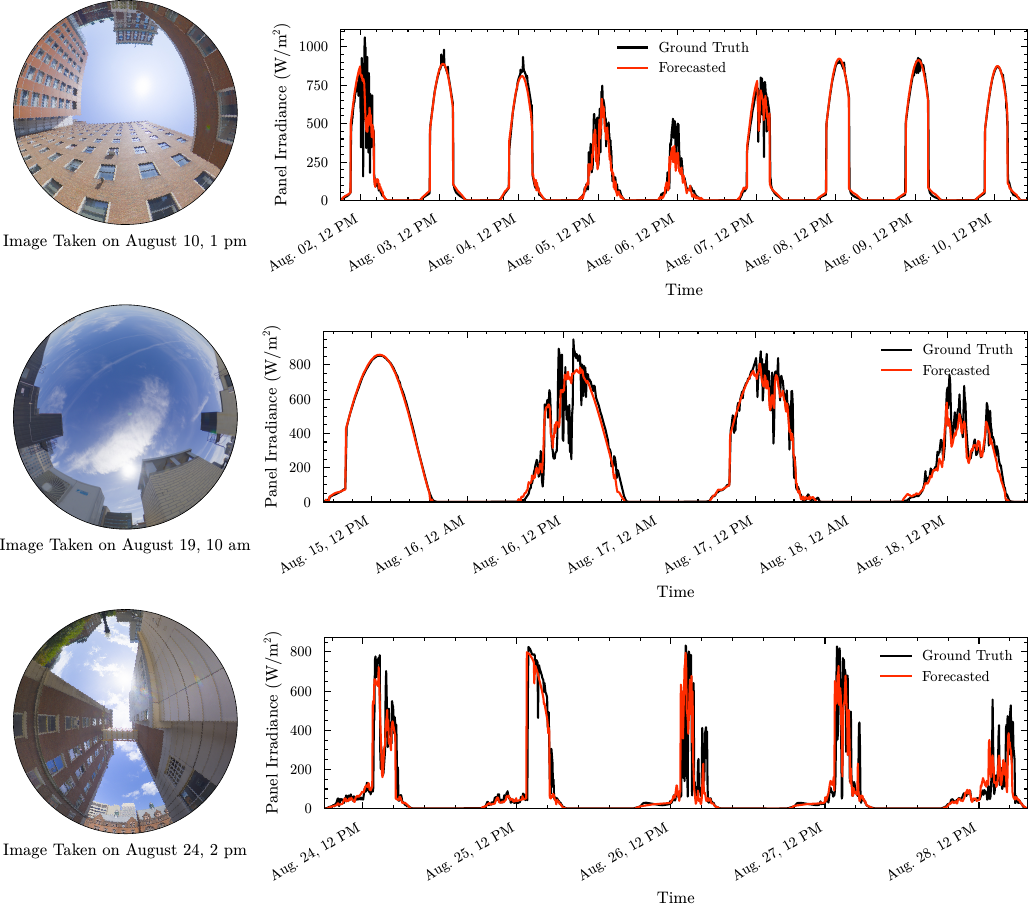}
    \caption{
        \textbf{Forecasted and ground truth irradiance over extended time horizons.} 
        Forecasted (red line) and ground truth (black line) panel irradiance across multiple days at three of the locations in \cref{fig:experiments}. See \cref{tab:experiment-results} for the sky conditions on each day and the errors in the forecasted daily irradiance.
    }
    \label{fig:app-experiments}
\end{figure*}

\Cref{fig:app-experiments} shows extended results for the forecasted (red line) and ground truth (black line) panel irradiance across multiple days at the four locations in \cref{fig:experiments}. The forecasted irradiance was computed using the single hemispherical image shown on the left. 

\subsection{Implementation Details of the Irradiance-Based Approach}
\label{app:irradiance-based-implementation-details}
We have implemented the irradiance-based approach following the calculation of the plane-of-array (POA) irradiance implemented by the System Advisory Model (SAM)~\cite{gilmanSAMPhotovoltaicModel2018}. Specifically, we compute panel irradiance as
\begin{equation}
    E_{total} = E_{DNI} \, A(\theta_s, \phi_s) \, (\vn \cdot \vs) + E_{diffuse} \cdot SVF + E_{ground},
\end{equation}
where $\theta_s$ and $\phi_s$ are the zenith and azimuth angles of the sun, respectively, $A(\theta_s, \phi_s)$ is the value of the sky aperture at the sun's location indicating if the sun is visible, $\vn$ is the normal vector of the panel, $\vs$ is the direction of the sun, $E_{diffuse}$ is the diffuse irradiance from the sky computed using the Perez transposition model~\cite{perezModelingDaylightAvailability1990}, and $SVF$ is the sky view factor. Following the implementation of SAM~\cite{gilmanSAMPhotovoltaicModel2018}, the ground irradiance is computed as
\begin{equation}
    E_{ground} = \rho \left( E_{DNI} \cos\theta_s + E_{DHI} \right) \left( \frac{1 - \cos\theta_p}{2} \right), \label{eq:Eground}
\end{equation}
where $\rho$ is the ground albedo (which we set to $0.2$), and $\theta_p$ is the zenith angle of the panel's normal vector. 

As noted in the paper, attenuating the diffuse irradiance by the sky view factor without considering which portion of the sky is visible to the panel can lead to errors in urban canyons. These errors are shown in \cref{tab:experiment-results} under ``Irradiance-Based Model 1,'' in which this approach often underestimates the irradiance due to the sky, leading to underestimates in the total panel irradiance.

Additionally, this model does not account for irradiance due to reflections from nearby buildings. In fact, for a panel pointing straight up (with $\theta_p = 0$), the model's estimated ground irradiance component is zero (see \cref{eq:Eground}). However, this is not the case in urban canyons in which scene irradiance accounts for, on average, $12\%$ of a panel's total annual irradiance (see \cref{sec:scene-irradiance}).

\subsection{Implementation Details of the 3D Model-Based Approach}
\label{app:extended-results-3d-model-comparison}
We have also compared the performance of forecasting a panel's irradiance using ray tracing within a 3D model. We used a model of the buildings in New York City derived from aerial imagery~\cite{nycofficeoftechnology&innovation3DBuildingModel} along with a digital elevation model derived from a LiDAR scan~\cite{nycofficeoftechnology&innovation1FootDigital} to construct a 3D model. Since this model does not include the material properties of the ground or buildings, we set the reflectance of all materials to be Lambertian. We selected a ground albedo of $0.2$ and a building wall and rooftop albedo of $0.3$. Then, we used a path tracer in Mitsuba~\cite{wenzeljakobMitsuba3Renderer2022} to render the illumination seen by the panel.

\begin{figure*}[t]
    \centering
    \includegraphics[width=\linewidth]{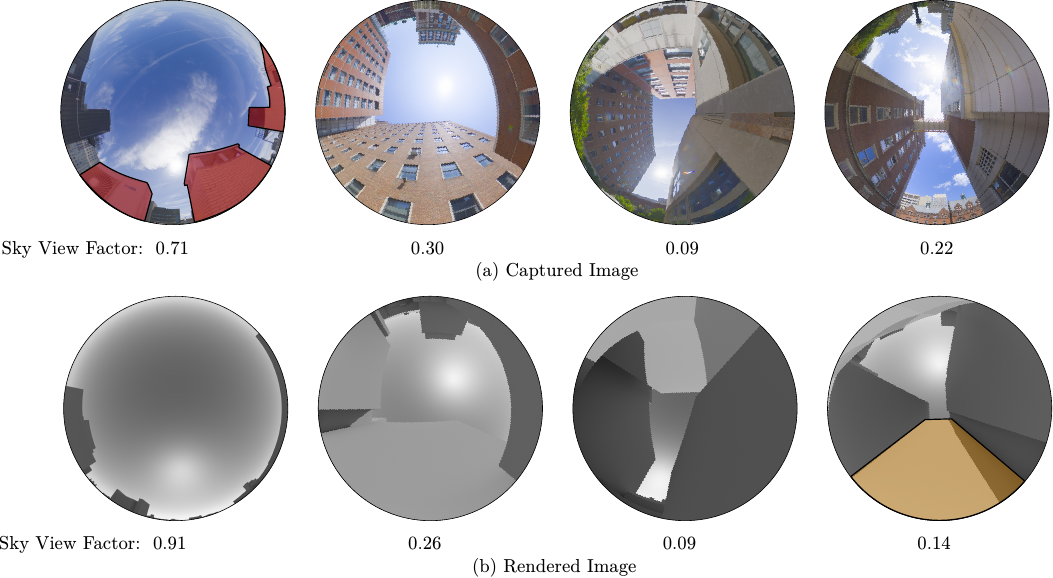}
    \caption{
        \textbf{Comparison of captured and rendered images for the real experiment locations.} 
        (a) Real image captured at each location in \cref{fig:experiments}. (b) Rendered image at the same location and time of day. As discussed in \cref{sec:comparison}, existing 3D models do not capture the small urban structures near a panel that obstruct its view of the sky. For example, the panel in the first column is placed near an HVAC unit, ventilation stack, and elevator shaft (highlighted in red), all of which are not captured by the 3D model. In addition, the 3D model in the final column erroneously contains an extra wall (highlighted in gold). The sky view factor listed below each image shows how these inaccuracies in the 3D model yield incorrect estimates of the panel's sky aperture.
    }
    \label{fig:supp-real-v-rendered}
\end{figure*}

\Cref{fig:supp-real-v-rendered} compares the real images captured at each location in \cref{fig:experiments} with the corresponding rendered image. Our approach of capturing a single image from the panel's location reveals all of the objects that obstruct a panel's view of the sky, whereas the 3D model does not accurately capture the surrounding environment.

%% file: sections/app06_panel_orientation.tex
\section{Determining Panel Orientation in the Camera Coordinate Frame}
\label{app:panel-orientation}
Consider a spherical camera placed near an existing panel as shown in \cref{fig:spherical}(a). The spherical camera captures two hemispherical images from opposing views, with the panel appearing in one of them (see \cref{fig:spherical}(b)). Given the image captured by the spherical camera, we seek to find the orientation of the panel in the camera coordinate frame. 

Finding the panel's orientation is equivalent to finding the 3D plane in which the panel resides. 
For simplicity, we present the approach using a perspective image, but this approach extends to any other image with a known projection model (including spherical images). We consider two different cases based on the number of the panel's corner that are visible in the image. In both cases, we assume the panel is rectangular (i.e.,~opposite edges on the panel's boundary are parallel).

\vspace{0.1in}
\noindent\textbf{Case 1: Four corners of the panel are visible in the image.}
In \cref{fig:spherical}(b), all four corners are visible in the image. These panel corners define the boundary edges of the panel. Since we assume the panel is rectangular, opposing boundary edges correspond to parallel scene lines. We use the vanishing point of the image of the parallel lines to determine their orientation in the camera coordinate frame~\cite{hartleyMultipleViewGeometry2003}. This leads to an algorithm to find the panel orientation:

\begin{enumerate}
    \item Use the location of the panel's four corners to compute the four lines of the panel's boundary.
    \item Compute the vanishing point of the two pairs of opposing (parallel) lines on the panel's boundary.
    \item Using the two vanishing points, compute two 3D rays passing through the camera center that are parallel to the panel's edges.
    \item The normal vector of the panel is given by the cross product of the two rays.
\end{enumerate}

\vspace{0.1in}
\noindent\textbf{Case 2: Two corners of the panel are visible in the image.}
The computation is simpler when just two corners of the panel are visible in the image:
\begin{enumerate}
    \item Compute the backprojected rays from the camera center to the two corners visible in the image.
    \item The normal vector of the panel is given by the cross product of the backprojected rays.
\end{enumerate}

In both cases, we set the sign of the normal vector such that the panel points above the horizon, since we can safely assume that a panel is never pointing toward the ground.

%% file: main.bbl
\begin{thebibliography}{10}
\expandafter\ifx\csname url\endcsname\relax
  \def\url#1{\texttt{#1}}\fi
\expandafter\ifx\csname urlprefix\endcsname\relax\def\urlprefix{URL }\fi
\expandafter\ifx\csname href\endcsname\relax
  \def\href#1#2{#2} \def\path#1{#1}\fi

\bibitem{kuhnReviewTechnologicalDesign2021}
T.~E. Kuhn, C.~Erban, M.~Heinrich, J.~Eisenlohr, F.~Ensslen, D.~H. Neuhaus,
  Review of technological design options for building integrated photovoltaics
  (bipv), Energy and Buildings 231 (2021) 110381.

\bibitem{shuklaComprehensiveReviewDesign2016}
A.~K. Shukla, K.~Sudhakar, P.~Baredar, A comprehensive review on design of
  building integrated photovoltaic system, Energy and Buildings 128 (2016)
  99--110.

\bibitem{traverseEmergenceHighlyTransparent2017}
C.~J. Traverse, R.~Pandey, M.~C. Barr, R.~R. Lunt, Emergence of highly
  transparent photovoltaics for distributed applications, Nature Energy 2~(11)
  (2017) 849--860.

\bibitem{solarwindowtechnologiesLiquidElectricity}
{Solar Window Technologies}, Liquidelectricity, https://www.solarwindow.com.

\bibitem{nextenergytechnologiesNEXTOPVTransparent}
{NEXT Energy Technologies}, Next opv transparent coatings,
  https://www.nextenergytech.com.

\bibitem{efacadeMitrex}
{eFacade}, Mitrex, https://www.mitrex.com.

\bibitem{newyorkindependentsystemoperator2024ReliabilityNeeds24}
{New York Independent System Operator}, 2024 reliability needs assessment
  ({RNA}), Tech. rep. (Nov. 24).

\bibitem{nationalrenewableenergylaboratorynrelSolarManufacturingCost}
{National Renewable Energy Laboratory (NREL)}, Solar manufacturing cost
  analysis,
  https://www.nrel.gov/solar/market-research-analysis/solar-manufacturing-cost.

\bibitem{nationalrenewableenergylaboratorynrelSolarInstalledSystem}
{National Renewable Energy Laboratory (NREL)}, Solar installed system cost
  analysis,
  https://www.nrel.gov/solar/market-research-analysis/solar-installed-system-cost.

\bibitem{oshaughnessyAddressingSoftCost2019}
E.~O'Shaughnessy, G.~F. Nemet, J.~Pless, R.~Margolis, Addressing the soft cost
  challenge in {U.S.} small-scale solar pv system pricing, Energy Policy 134
  (2019) 110956.

\bibitem{klotzMinimalSensingOrienting2025}
J.~Klotz, S.~K. Nayar, Minimal sensing for orienting a solar panel, Solar
  Energy 300 (2025) 113833.

\bibitem{jakubiecMethodPredictingCitywide2013}
J.~A. Jakubiec, C.~F. Reinhart, A method for predicting city-wide electricity
  gains from photovoltaic panels based on {LiDAR} and {GIS} data combined with
  hourly daysim simulations, Solar Energy 93 (2013) 127--143.

\bibitem{freitasModellingSolarPotential2015}
S.~Freitas, C.~Catita, P.~Redweik, M.~Brito, Modelling solar potential in the
  urban environment: State-of-the-art review, Renewable and Sustainable Energy
  Reviews 41 (2015) 915--931.

\bibitem{mardaljevicIrradiationMappingComplex2003}
J.~Mardaljevic, M.~Rylatt, Irradiation mapping of complex urban environments:
  An image-based approach, Energy and Buildings 35~(1) (2003) 27--35.

\bibitem{zhuEffectUrbanMorphology2020}
R.~Zhu, M.~S. Wong, L.~You, P.~Santi, J.~Nichol, H.~C. Ho, L.~Lu, C.~Ratti, The
  effect of urban morphology on the solar capacity of three-dimensional cities,
  Renewable Energy 153 (2020) 1111--1126.

\bibitem{robinsonSolarRadiationModelling2004}
D.~Robinson, A.~Stone, Solar radiation modelling in the urban context, Solar
  Energy 77~(3) (2004) 295--309.

\bibitem{robinsonSUNtoolNewModelling2007}
D.~Robinson, N.~Campbell, W.~Gaiser, K.~Kabel, A.~{Le-Mouel}, N.~Morel,
  J.~Page, S.~Stankovic, A.~Stone, {SUNtool} -- a new modelling paradigm for
  simulating and optimising urban sustainability, Solar Energy 81~(9) (2007)
  1196--1211.

\bibitem{compagnonSolarDaylightAvailability2004}
R.~Compagnon, Solar and daylight availability in the urban fabric, Energy and
  Buildings 36~(4) (2004) 321--328.

\bibitem{jakicaStateoftheartReviewSolar2018}
N.~Jakica, State-of-the-art review of solar design tools and methods for
  assessing daylighting and solar potential for building-integrated
  photovoltaics, Renewable and Sustainable Energy Reviews 81 (2018) 1296--1328.

\bibitem{kosmopoulosRayTracingModelingUrban2024}
P.~Kosmopoulos, H.~Dhake, D.~Kartoudi, A.~Tsavalos, P.~Koutsantoni,
  A.~Katranitsas, N.~Lavdakis, E.~Mengou, Y.~Kashyap, Ray-tracing modeling for
  urban photovoltaic energy planning and management, Applied Energy 369 (2024)
  123516.

\bibitem{anSolarEnergyPotential2023}
Y.~An, T.~Chen, L.~Shi, C.~K. Heng, J.~Fan, Solar energy potential using
  {GIS}-based urban residential environmental data: A case study of shenzhen,
  china, Sustainable Cities and Society 93 (2023) 104547.

\bibitem{googleProjectSunroof}
{Google}, Project sunroof, https://sunroof.withgoogle.com.

\bibitem{martinez-rubioEvaluatingSolarIrradiance2016}
A.~{Mart{\'i}nez-Rubio}, F.~{Sanz-Adan}, J.~{Santamar{\'i}a-Pe{\~n}a},
  A.~Mart{\'i}nez, Evaluating solar irradiance over facades in high building
  cities, based on {LiDAR} technology, Applied Energy 183 (2016) 133--147.

\bibitem{chengSolarEnergyPotential2020}
L.~Cheng, F.~Zhang, S.~Li, J.~Mao, H.~Xu, W.~Ju, X.~Liu, J.~Wu, K.~Min,
  X.~Zhang, M.~Li, Solar energy potential of urban buildings in 10 cities of
  china, Energy 196 (2020) 117038.

\bibitem{andresTimevaryingRayTracing2023}
C.~Andres, C.~Ruben, G.~David, B.~Pavel, M.~Patrizio, Z.~Miro, I.~Olindo,
  Time-varying, ray tracing irradiance simulation approach for photovoltaic
  systems in complex scenarios with decoupled geometry, optical properties and
  illumination conditions, Progress in Photovoltaics: Research and Applications
  31~(2) (2023) 134--148.

\bibitem{calcabriniSimplifiedSkylinebasedMethod2019}
A.~Calcabrini, H.~Ziar, O.~Isabella, M.~Zeman, A simplified skyline-based
  method for estimating the annual solar energy potential in urban
  environments, Nature Energy 4~(3) (2019) 206--215.

\bibitem{meteonormHoricatcher}
{Meteonorm},
  \href{https://mn8.meteonorm.com/en/product/horicatcher}{Horicatcher}.
\newline\urlprefix\url{https://mn8.meteonorm.com/en/product/horicatcher}

\bibitem{zhangDeepPhotovoltaicNowcasting2018}
J.~Zhang, R.~Verschae, S.~Nobuhara, J.-F. Lalonde, Deep photovoltaic
  nowcasting, Solar Energy 176 (2018) 267--276.

\bibitem{julianComputationalImagingLongTerm2025}
L.~K. Julian, H.~Lee, S.~Kar, A.~C. Sankaranarayanan, Computational imaging for
  long-term prediction of solar irradiance, IEEE Transactions on Pattern
  Analysis and Machine Intelligence 47~(9) (2025) 7182--7193.

\bibitem{antonanzasReviewPhotovoltaicPower2016}
J.~Antonanzas, N.~Osorio, R.~Escobar, R.~Urraca, F.~{Martinez-de-Pison},
  F.~{Antonanzas-Torres}, Review of photovoltaic power forecasting, Solar
  Energy 136 (2016) 78--111.

\bibitem{gilmanSAMPhotovoltaicModel2018}
P.~Gilman, N.~A. DiOrio, J.~M. Freeman, S.~Janzou, A.~Dobos, D.~Ryberg, Sam
  photovoltaic model technical reference update, Tech. Rep.
  NREL/TP--6A20-67399, 1429291 (Mar. 2018).

\bibitem{michalskyAstronomicalAlmanacAlgorithm1988}
J.~J. Michalsky, {\emph{The Astronomical Almanac}}'s algorithm for approximate
  solar position (1950--2050), Solar Energy 40~(3) (1988) 227--235.

\bibitem{meeusAstronomicalAlgorithms1991}
J.~H. Meeus, Astronomical Algorithms, Willmann-Bell, Incorporated, Richmond,
  VA, 1991.

\bibitem{lalondeEstimatingNaturalIllumination2012}
J.-F. Lalonde, A.~A. Efros, S.~G. Narasimhan, Estimating the natural
  illumination conditions from a single outdoor image, International Journal of
  Computer Vision 98~(2) (2012) 123--145.

\bibitem{workmanHorizonLinesWild2016}
S.~Workman, M.~Zhai, N.~Jacobs, Horizon lines in the wild, in: Procedings of
  the British Machine Vision Conference 2016, British Machine Vision
  Association, York, UK, 2016, pp. 20.1--20.12.

\bibitem{xianUprightNetGeometryAwareCamera2019}
W.~Xian, Z.~Li, N.~Snavely, M.~Fisher, J.~Eisenman, E.~Shechtman, Uprightnet:
  Geometry-aware camera orientation estimation from single images, in: 2019
  IEEE/CVF International Conference on Computer Vision (ICCV), IEEE, Seoul,
  Korea (South), 2019, pp. 9973--9982.

\bibitem{hold-geoffroyPerceptualMeasureDeep2018}
Y.~{Hold-Geoffroy}, K.~Sunkavalli, J.~Eisenmann, M.~Fisher, E.~Gambaretto,
  S.~Hadap, J.-F. Lalonde, A perceptual measure for deep single image camera
  calibration, in: 2018 IEEE/CVF Conference on Computer Vision and Pattern
  Recognition, IEEE, Salt Lake City, UT, USA, 2018, pp. 2354--2363.

\bibitem{wuTinyViTFastPretraining2022}
K.~Wu, J.~Zhang, H.~Peng, M.~Liu, B.~Xiao, J.~Fu, L.~Yuan, Tinyvit: Fast
  pretraining distillation for small vision transformers, in: S.~Avidan,
  G.~Brostow, M.~Ciss{\'e}, G.~M. Farinella, T.~Hassner (Eds.), Computer Vision
  -- ECCV 2022, Vol. 13681, Springer Nature Switzerland, Cham, 2022, pp.
  68--85.

\bibitem{kirillovSegmentAnything2023}
A.~Kirillov, E.~Mintun, N.~Ravi, H.~Mao, C.~Rolland, L.~Gustafson, T.~Xiao,
  S.~Whitehead, A.~C. Berg, W.-Y. Lo, P.~Doll{\'a}r, R.~Girshick, Segment
  anything, in: 2023 IEEE/CVF International Conference on Computer Vision
  (ICCV), IEEE, Paris, France, 2023, pp. 3992--4003.

\bibitem{perezAllweatherModelSky1993}
R.~Perez, R.~Seals, J.~Michalsky, All-weather model for sky luminance
  distribution---preliminary configuration and validation, Solar Energy 50~(3)
  (1993) 235--245.

\bibitem{wenzeljakobMitsuba3Renderer2022}
{Wenzel Jakob}, {S\'ebastien Speierer}, {Nicolas Roussel}, {Merlin
  Nimier-David}, {Delio Vicini}, {Tizian Zeltner}, {Baptiste Nicolet}, {Miguel
  Crespo}, {Vincent Leroy}, {Ziyi Zhang}, Mitsuba 3 renderer (2022).

\bibitem{duffieSolarEngineeringThermal2006}
J.~A. Duffie, W.~A. Beckman, Solar Engineering of Thermal Processes, John Wiley
  \& Sons, Inc., Hoboken, New Jersey, 2006.

\bibitem{hartleyMultipleViewGeometry2003}
R.~Hartley, A.~Zisserman, Multiple View Geometry in Computer Vision, 2nd
  Edition, Cambridge University Press, 2003.

\bibitem{hosekAnalyticModelFull2012}
L.~Hosek, A.~Wilkie, An analytic model for full spectral sky-dome radiance, ACM
  Transactions on Graphics 31~(4) (2012) 1--9.

\bibitem{ineichenNewAirmassIndependent2002}
P.~Ineichen, R.~Perez, A new airmass independent formulation for the linke
  turbidity coefficient, Solar Energy 73~(3) (2002) 151--157.

\bibitem{perezNewOperationalModel2002}
R.~Perez, P.~Ineichen, K.~Moore, M.~Kmiecik, C.~Chain, R.~George, F.~Vignola, A
  new operational model for satellite-derived-irradiances: Description and
  validation (2002).

\bibitem{loshchilovDecoupledWeightDecay2019}
I.~Loshchilov, F.~Hutter, Decoupled weight decay regularization, in:
  International Conference on Learning Representations, 2019.

\bibitem{JapaneseCityMidtown}
Japanese city - midtown environment,
  https://www.cgtrader.com/3d-models/exterior/cityscape/japanese-city-midtown-environment.

\bibitem{brightSolcastValidationSatellitederived2019}
J.~M. Bright, Solcast: Validation of a satellite-derived solar irradiance
  dataset, Solar Energy 189 (2019) 435--449.

\bibitem{yangWorldwideValidation82020}
D.~Yang, J.~M. Bright, Worldwide validation of 8 satellite-derived and
  reanalysis solar radiation products: A preliminary evaluation and overall
  metrics for hourly data over 27 years, Solar Energy 210 (2020) 3--19.

\bibitem{perezModelingDaylightAvailability1990}
R.~Perez, P.~Ineichen, R.~Seals, J.~Michalsky, R.~Stewart, Modeling daylight
  availability and irradiance components from direct and global irradiance,
  Solar Energy 44~(5) (1990) 271--289.

\bibitem{nycofficeoftechnology&innovation3DBuildingModel}
{NYC Office of Technology \& Innovation}, 3-{D} building model,
  https://data.cityofnewyork.us/City-Government/3-D-Building-Model/tnru-abg2.

\bibitem{nycofficeoftechnology&innovation1FootDigital}
{NYC Office of Technology \& Innovation}, 1 foot digital elevation model (dem),
  https://data.cityofnewyork.us/City-Government/1-foot-Digital-Elevation-Model-DEM-/dpc8-z3jc.

\end{thebibliography}
